\documentclass[11pt]{article}

\usepackage{amsmath,amsfonts,bm}

\def\eqref#1{equation~\ref{#1}}

\def\1{\bm{1}}

\DeclareMathAlphabet{\mathsfit}{\encodingdefault}{\sfdefault}{m}{sl}
\SetMathAlphabet{\mathsfit}{bold}{\encodingdefault}{\sfdefault}{bx}{n}

\usepackage[numbers, compress]{natbib}
\usepackage[margin=1.1in]{geometry}
\usepackage{hyperref}
\usepackage{url}
\usepackage{amsmath}
\usepackage{amsfonts}
\usepackage{amssymb}
\usepackage{multicol, multirow}
\usepackage{booktabs}
\usepackage{subcaption}
\usepackage[table]{xcolor} 
\definecolor{RowGray}{gray}{0.90} 
\usepackage{graphicx}

\title{(Token-Level) \textbf{InfoRMIA}: 
Stronger Membership Inference and Memorization Assessment for LLMs}

\author{Jiashu Tao$^\dag$, and Reza Shokri$^\dag$$^\ddag$\\
	$^\dag$ National University of Singapore, 
	$^\ddag$ Google Research \\ 
	\texttt{\{jiashut, reza\}@comp.nus.edu.sg}\\
}

\begin{document}

\maketitle

\begin{abstract}
Machine learning models are known to leak sensitive information, as they inevitably memorize (parts of) their training data. More alarmingly, large language models (LLMs) are now trained on nearly all available data, which amplifies the magnitude of information leakage and raises serious privacy risks. Hence, it is more crucial than ever to quantify privacy risk before the release of LLMs. The standard method to quantify privacy is via membership inference attacks, where the state-of-the-art approach is the Robust Membership Inference Attack (RMIA). In this paper, we present InfoRMIA, a principled information-theoretic formulation of membership inference. Our method consistently outperforms RMIA across benchmarks while also offering improved computational efficiency.

In the second part of the paper, we identify the limitations of treating sequence-level membership inference as the gold standard for measuring leakage. We propose a new perspective for studying membership and memorization in LLMs: token-level signals and analyses. We show that a simple token-based InfoRMIA can pinpoint which tokens are memorized within generated outputs, thereby localizing leakage from the sequence level down to individual tokens, while achieving stronger sequence-level inference power on LLMs. This new scope rethinks privacy in LLMs and can lead to more targeted mitigation, such as exact unlearning.
\end{abstract}
\section{Introduction}
In the past decade, researchers have shown that machine learning (ML) models inevitably memorize parts of their training data~\citep{feldman2020does, feldman2020neural}. Memorized data, once identified and extracted, can pose a severe privacy risk. It is increasingly concerning as the contemporary, easily accessible large language models (LLMs) are trained on datasets so large that we are running out of training data~\citep{villalobos2024position}. These LLMs have seen nearly all data generated by humans. Even limited memorization by them can translate into significant privacy risks.

The current standard for quantifying privacy is membership inference attacks (MIAs)~\citep{shokri2017membership}, where the attacker or privacy auditor aims to determine if a given data sample was part of the target model’s training set. A stronger attack means the attacker can more accurately distinguish members (training data) from non-members, implying that the target model leaks more of its training data. This ability to separate members from non-members not only signals privacy risk but also raises the possibility of training data reconstruction. It is also closely linked to memorization, as it is the root cause of successful MIAs. Hence, MIAs are widely regarded as the backbone of ML privacy research. The state-of-the-art (SOTA) MIA is the Robust Membership Inference Attack (RMIA)~\citep{zarifzadeh2024low}, but its dependence on a separate population dataset, whose size scales linearly with the training set, could be a potential concern, especially for LLMs.

In the first part of the paper, we thoroughly analyze RMIA, from its formulation to signal computation, and propose a more principled statistical test by casting RMIA’s setup as a composite hypothesis testing problem. Our approach can also be interpreted through information theory, where we quantify dominance over population data in bits rather than in sample counts. This transforms the attack signal from discrete to continuous, eliminating the sensitivity on the population dataset size. We observe that our new attack, InfoRMIA, consistently outperforms RMIA on tabular, image, and text datasets, while requiring far fewer population samples. Thus, InfoRMIA is a lower-cost, higher-power membership inference attack and establishes a new SOTA.

\begin{figure*}[!t]
    \centering
    \begin{subfigure}[b]{0.48\textwidth}
        \centering
        \includegraphics[width=0.9\textwidth]{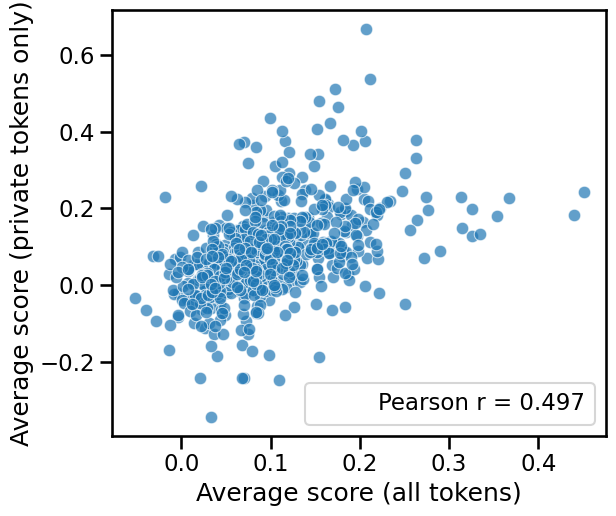}
        \caption{The average membership scores of sequences and their private tokens are not strongly correlated. }
        \label{fig:seq_score_vs_token_score}
    \end{subfigure}%
    \hfill
    \begin{subfigure}[b]{0.48\textwidth}
        \centering
        \includegraphics[width=0.95\textwidth]{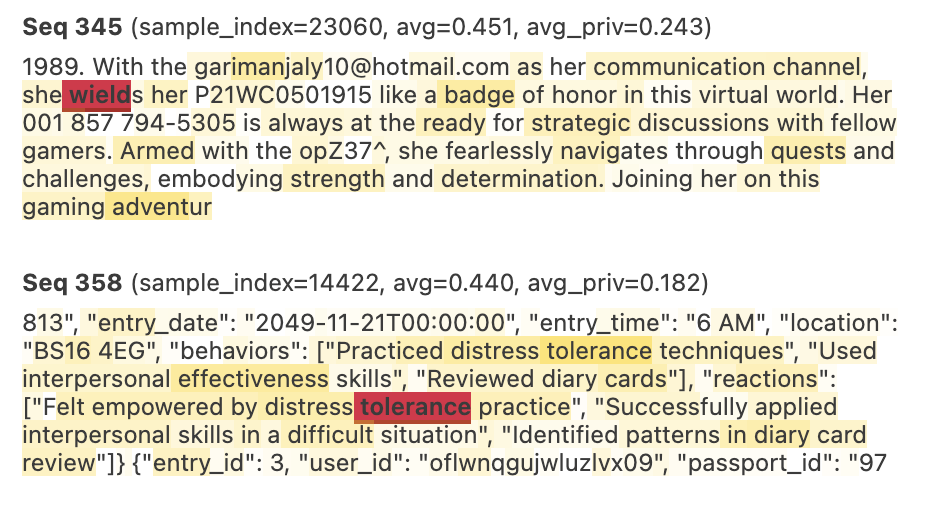}
        \caption{Heatmaps of token-based membership scores on the input. The two most memorized sequences, identified by sequence-level membership scores, mainly memorize non-private tokens.}
        \label{fig:viz}
    \end{subfigure}
    \caption{Sequence-level membership inference may not accurately identify private information leakage, which is conveyed by private tokens only.}
    \label{fig:seq_mia_not_useful}
    \vspace{-0.5cm}
\end{figure*}

Although MIAs are the gold standard in quantifying privacy, they must follow a strict setup defined by the membership inference game~\citep{yeom2018privacy, ye2022enhanced, zarifzadeh2024low}, which falls short in quantifying true information leakage~\citep{tao2025range}, especially for LLMs. The current privacy quantification setup for LLMs is almost identical to those designed for MLPs and CNNs: perform MIA on a set of member and non-member sequences. However, transformers are sequential models that generate predictions token by token. Assigning a single membership label to an entire sequence, list of outputs rather than a single one, compresses rich token-based information into a single bit, losing granularity and analogous to a lossy compression (Figure~\ref{fig:seq_score_vs_token_score}).

To address this issue, in the second part, we propose a token-level MIA framework to better quantify memorization and information leakage from LLMs with finer granularity and more meaningful analysis. There are three main reasons for doing it on the token level instead of the sequence level. First, since each token completion is one prediction step, analyzing leakage at the token level naturally aligns with model behavior and the definition of MIA. Second, sequence-level metrics are aggregated from token-level ones, making them less semantically meaningful for privacy assessment. For example, memorized facts may yield high sequence-based membership scores despite not leaking private information. Third, we argue that private information in a sentence is usually contained in a few words/tokens. Measuring the average memorization of the entire sequence mainly with common words leads to inaccurate privacy assessment (Figure~\ref{fig:viz}). Token-level analysis can narrow the focus to truly sensitive components, enabling more accurate privacy quantification. By pinpointing the information leakage to individual tokens and words, we can potentially protect privacy more effectively by performing targeted machine unlearning, which would prevent unlearning useful information from non-private texts, while surgically removing the memorization of private information.

In summary, this paper makes two main contributions: (1) we propose InfoRMIA, a principled and efficient improvement over RMIA that achieves new SOTA performance; and (2) we introduce a token-level privacy assessment framework, which offers finer-grained insights into memorization and leakage in LLMs, while achieving strong membership inference capabilities.

\section{Related Work}
\subsection{Membership Inference}
Membership inference~\citep{shokri2017membership} is a class of inference attacks that aims to determine if any given point query $x$ is part of the training set of a machine learning model $\theta$. Since its inception, there has been significant progress in the inference strategies. \citet{shokri2017membership} trained shadow models to predict the membership label directly. \citet{yeom2018privacy} proposed a simpler approach that uses the loss values as the membership signal. 
To achieve higher inference accuracy, researchers have proposed multiple reference model-based membership inference tests to calibrate the raw signal on the target query. \citet{ye2022enhanced} trained a set of reference models on the population dataset, and counted the number of them with lower probability on the target $x$. \citet{carlini2022membership} constructed reference models that train with or without the target point to simulate two distributions of model outputs on the target $x$: the IN and OUT distributions. Assuming Gaussians, the attacker computes the likelihood ratios under the two distributions as the membership signal. The state-of-the-art attack, RMIA~\citep{zarifzadeh2024low}, improves further upon reference model-based attacks by counting how many similar data points each test point dominates.

Membership inference techniques on CNNs and MLPs can be adapted to work on LLMs, where the goal is to predict if any given \emph{text sequence} is part of the training dataset. Due to the high computation cost to train reference models, LLM-specific MIAs tend to be reference model-free. \citet{carlini2021extracting} used entropy, or more easily, zlib~\citep{gailly2004zlib} compression, to calibrate sequence-based membership likelihood. \citet{mattern2023membership} compared the perplexity gap between the target and neighboring sequences, while \citep{shi2024detecting} looked at the tokens with the smallest probability. \citet{duan2024membership} published a benchmark, MIMIR, to evaluate LLM MIAs and found that all of these methods perform poorly on pretrained LLMs. \citet{zhang2024pretraining} and \citet{zhang2025mink} improved upon the methods in MIMIR by incorporating additionalx calibration.

\subsection{Memorization}
Memorization of machine learning models is defined in a leave-one-out fashion by \citet{feldman2020does}. Due to its prohibitively high computation cost on LLMs, many alternative definitions have been proposed. So far, verbatim memorization~\citep{carlini2021extracting, carlini2023quantifying}, which means the output sequence \textbf{exactly} matches one of the training sequences, is the most popular notion. If the sequence is generated verbatim when conditioned on a given prompt, the memorization term is called discoverable~\citep{carlini2023quantifying, nasr2023scalable} or extractable memorization~\citep{nasr2023scalable}, depending on whether the prompt is crafted by an adversary. \citet{hayes2025measuring} introduced their probabilistic variations, considering the stochastic nature of LLMs.  There are other memorization notions such as $k$-eidetic~\citep{carlini2021extracting} and counterfactual memorization~\citep{zhang2023counterfactual}.

\section{Improving RMIA with an Information-Theoretic Inspired Test Statistic}
We introduce an improved version of RMIA~\citep{zarifzadeh2024low}, which we call information-theoretic RMIA (InfoRMIA). This new attack is consistently stronger than the original RMIA across all datasets and thus establishes a new state-of-the-art.
\subsection{The Original Robust Membership Inference Attacks (RMIA)}
The original test statistic of RMIA can be written as
\begin{equation} \label{eqn:rmia_original}
    p_z\left(\frac{p(\theta|x)}{p(\theta|z)}\geq \gamma\right) = \frac{1}{|Z|}\sum_{z \in Z} \mathbb{I}\left(\frac{p(x|\theta)}{p(x)}/ \frac{p(z|\theta)}{p(z)}\geq \gamma\right),
\end{equation} where $\theta$ is the target model, $x$ is the target query, $z$ is population data, and $\gamma \geq 1$ is the threshold. It counts the proportion of ``similar" data the target $x$ dominates. The score is a discrete value, whose precision depends on $|Z|$. More details of RMIA can be found in Appendix~\ref{app:rmia}.

\subsection{Info-Theoretic RMIA (InfoRMIA)}
Instead of counting how many population \emph{data} the target point dominates, we measure how many \emph{bits} the target point saves in explaining the target model relative to the population data in expectation.

That is, we want to measure 
\begin{equation} \label{eqn:bits_rmia}
    \mathbb{E}_z \left[-\log p(\theta|z) \right] - (-\log p(\theta|x))=\log p(\theta|x) - \mathbb{E}_z \log p(\theta|z)
\end{equation}

By applying the same Bayesian decomposition in~\cite{zarifzadeh2024low} and some basic manipulations, we can obtain the following equivalent formulation of our new test statistic:
\begin{equation} \label{eqn:new_rmia}
\begin{aligned} 
\text{Test Statistic} &= \sum_z p(z) \log\left(\frac{p(\theta | x)}{p(\theta | z)}\right) = \sum_z p(z) \log\left(\frac{p(x|\theta)p(z)}{p(z|\theta)p(x)}\right) \\ 
                      &= \log\left(\frac{p(x|\theta)}{p(x)}\right) + \sum_z p(z)\log\left(\frac{p(z)}{p(z|\theta)}\right) \\ 
                      &= \log\left(\frac{p(x|\theta)}{p(x)}\right) + D_{\text{KL}}\left(p(z) \ || \ p(z|\theta)\right) 
\end{aligned}
\end{equation}

Note that the formulation is only valid when $\sum_z p(z)=1$\footnote{Actually, when attacking \textbf{one single fixed} model with a \textbf{fixed} population dataset, the attack performance is unchanged even if $\sum_z p(z) \neq 1$. This is because the test statistics would be reduced to $\sum_z p(z) \log p(\theta|x)=C \log p(\theta|x)$, where $C=\sum_z p(z) > 0$ is a constant. Hence, the test statistic would preserve the same total order among all $x$'s log likelihood values.}. Hence, for an empirical or approximated (in RMIA's case) $\Tilde{p}(z)$, we need to normalize it to $\hat{p}(z)=\Tilde{p}(z)/\sum_z \Tilde{p}(z)$. Similarly, for the last step to hold, we require that $\sum_z p(z|\theta)=1$.  For simplicity, we use $p(z)$ and $p(z|\theta)$ in the rest of the paper to denote the normalized distribution of population data $z$. As this new test statistic is inspired by information theory, we refer to it as \emph{InfoRMIA}. 

\paragraph{Interpretation of the test statistic} It is interesting that the test statistic has two parts:
\begin{enumerate}
    \item $\log\left(\frac{p(x|\theta)}{p(x)}\right)$, which measures the amount of information gain in explaining $x$ given a model $\theta$. This can be seen as the memorization of $x$ by model $\theta$. 
    \item $D_{\text{KL}}\left(p(z) || p(z|\theta)\right)$, which captures the distributional differences between the base probabilistic distribution of $z$ and that conditioned on model $\theta$. This is reminiscent of generalization analysis, as it reflects the changes in model's predictive performance on $z$'s.
\end{enumerate}

\subsection{Why is InfoRMIA Better}
Both test statistics of the original and InfoRMIA are principled approaches to solve the same hypothesis testing problem derived from the same membership inference game \citep{yeom2018privacy, ye2022enhanced, carlini2022membership, zarifzadeh2024low}.
\begin{equation} \label{eqn:hypothesis}
\begin{aligned}
    &H_0: \text{The target model } \theta \text{ is trained with one of the data } z \in Z, \\
    &H_1: \text{The target model } \theta \text{ is trained with the given } x.
\end{aligned}
\end{equation}

The original RMIA (Equation~\ref{eqn:rmia_original}) performs multiple pairwise tests between $H_1$ and each null hypothesis. Each test requires a threshold $\gamma$. The final score is the proportion of null hypotheses rejected in all the pairwise tests. As mentioned before, this score is inherently discrete, with granularity determined by $|Z|$ and increments of $\frac{1}{|Z|}$.

InfoRMIA does not perform multiple pairwise tests. Instead, it opts for a more systematic approach. Similar to what \citet{tao2025range} observed, the scenario described by Equation~\ref{eqn:hypothesis} is a composite hypothesis testing problem. One of the principled solutions is to use Bayes Factor~\citep{tao2025range, jeffreys1939theory}, where we compute the expected log likelihood of the composite null hypothesis by $ \mathbb{E}_z \left[\log p(\theta|z) \right]$. Now it becomes clear that InfoRMIA's test statistic  (Equation~\ref{eqn:bits_rmia}) corresponds to the log of the likelihood ratio when using the Bayes Factor.

Apart from \textbf{using a more accurate and established test}, InfoRMIA also supersedes the original RMIA by using a \textbf{continuous test statistic} (See Equation~\ref {eqn:bits_rmia},  ~\ref{eqn:new_rmia}). This results in significantly higher precision in the membership score and also eliminates the need for the hyperparameter $\gamma$. Since the granularity of the score is no longer dictated by the size of $Z$, InfoRMIA is \textbf{much less dependent on a large $Z$}, significantly reducing computational overhead when $|Z|$ is fixed and lowering complexity by a constant factor. Experiment results in Section~\ref{sec:experiments} validate these improvements.
\section{Token-Level InfoRMIA for Attacking LLMs}
We have now justified why InfoRMIA surpasses the original RMIA and becomes the new SOTA attack\footnote{Although the authors in \url{https://arxiv.org/abs/2505.18773} found that LiRA~\citep{carlini2022membership} performs better than RMIA with a large number of reference models for LLMs, we found that their RMIA implementation deviates from the RMIA paper, which can affect RMIA's performance. But even now, the two attacks are within standard deviations with many reference models. With limited reference models, RMIA is better. }. We now propose our token-level framework where we can pinpoint information leakage and more truthfully estimate privacy risks with token-level InfoRMIA.

\subsection{From Sequences to Tokens}
So far, membership inference and privacy risks for LLMs have been defined on the sequence level, i.e., whether a given sequence is a member. The majority of the LLM MIAs aim to compute a score on each sequence~\citep{carlini2021extracting, mattern2023membership} based on its perplexity. However, delving into the mechanisms of LLMs, we can quickly realize that a sequence is not one single output, but an ordered list of outputs. For example, given training sequence is $\mathbf{x}=\{x_1x_2\dots x_k\}$, the LLM $\theta$ optimizes the losses $\ell(x_2, \theta(x_1)), \ell(x_3, \theta(x_1x_2)), \dots, \ell(x_k, \theta(x_1\dots x_{k-1}))$. Each sequence is more than one training sample; it resembles a dataset containing $k-1$ training (subsequence, label) pairs. To properly adapt existing MIAs to LLMs, we should treat each token generation step, which ``labels" each ``prefixal" subsequence (subsequences from the start), as one prediction and compute its membership likelihood. In this way, for any sequence of length $k$, the LLM goes through $k-1$ prediction steps, and we should obtain $k-1$ membership scores. In comparison, the existing framework only computes a single membership signal for each sequence, which is a highly compressed signal, losing rich information at each token position.

The token-level framework also provides a more realistic privacy notion for LLMs. Many researchers have pointed out that the current privacy definition via membership inference is too strict and not comprehensive enough, especially for language data, as it only considers \emph{exact} matches as privacy concerns~\citep{tao2025range, duan2024membership}. We believe that the privacy risk of a text sequence primarily resides in the tokens carrying the sensitive information. From an information-theoretic point of view, the total private information in bits can be computed by
\begin{equation} \label{eqn:privacy_bits}
    \text{PrivBits}=\sum_{x \in V_{priv}} -\log p(x) < \sum_x - \log p(x),
\end{equation}
where $V_{priv}$ is the set of all privacy concerning tokens in the data. From the inequality, it is obvious that the existing privacy notion is treating all tokens in the member sequence as private, leading to inflated membership scores in evaluation. In this process, the true information leakage can be diluted or overshadowed on the sequence level (Figure~\ref{fig:top_10_token}), especially in long texts and documents. This masks the signals from the truly private tokens and leads to inaccurate auditing results (since we are evaluating the upper bounds). Moreover, a sequence-based analysis framework also fails to pinpoint the source of true leakage. This affects downstream tasks like unlearning: we cannot make the model forget the sensitive information, but rather entire documents that may contain useful general semantic knowledge. 

With a token-level framework, users can compute leakage via every token completion. They can then easily visualize what tokens are memorized outputs and check if they are sensitive~(Figure~\ref{fig:viz}). For auditors who know where the personally identifiable information (PII) is, they can also choose to directly check the model's memorization extent on the corresponding tokens. We build this interface and will explain in detail in Section~\ref{sec:viz}.

\subsection{Token-Level InfoRMIA}
Our token-level framework relies on a MIA that can operate on the token-level. We propose to conduct InfoRMIA (Equation~\ref{eqn:new_rmia}) token by token, treating all tokens $x$ as labels for their respective prefixal substrings and compute a token-based score. In addition, we no longer curate a separate population dataset $Z$. Instead, we treat all possible tokens in the vocabulary other than the ground-truth $x$ as $z$. 
In this way, we have a data-dependent $Z$ that removes the high cost of curating and computing on an independent population dataset. 

We want to emphasize that since $p(x|\theta)+\sum_{z\in Z} p(z|\theta) = \sum_{z \in V} p(z|\theta)=1$ and $\sum_{z \in V} p(z)=\sum_{z \in V} \text{Avg}_{\theta_\text{ref}}p(z|\theta_{\text{ref}})=\text{Avg}_{\theta_\text{ref}}\sum_{z \in V} p(z|\theta_{\text{ref}})=1$, we can have an equivalent formulation that does not require normalization (full derivation in Equation~\ref{eqn:token_info_rmia_deri}):
\begin{align}
  &  \sum_{z\in Z} p(z) \log\left(\frac{p(\theta | x)}{p(\theta | z)}\right) \\ 
  =& \sum_{z\in Z} p(z) \log\left(\frac{p(x|\theta)p(z)}{p(z|\theta)p(x)}\right)+ p(x) \log\left(\frac{p(x|\theta)p(x)}{p(x|\theta)p(x)}\right)  \\
 =&  \sum_{z \in V} p(z) \log\left(\frac{p(x|\theta)}{p(x)}\right) + \sum_{z \in V} p(z) \log\left(\frac{p(z)}{p(z|\theta)}\right)\\
 =& \log\left(\frac{p(x|\theta)}{p(x)}\right) + D_{\text{KL}}\left(p(z) \ || \ p(z|\theta)\right) \label{eqn:token_info_rmia_kl}
\end{align}
Equation~\ref{eqn:token_info_rmia_kl} serves as an alternative form of our test statistic in Equation~\ref{eqn:new_rmia} that works with normalized probabilities, where we can include $x$ in our $Z$ and compute the KL divergence on all possible token choices, without removing the ground truth token and gathering the remaining logits. 

\subsection{From Token-Level to Sequence-Level MIAs}
The reigning privacy auditing and evaluation framework is on the sequence level. Here, we describe how to use token-level MIAs to perform sequence-level MIAs. This is \textbf{not the optimal way to evaluate token-level MIAs} and more like a \textbf{proof of concpet}, but our results in Section~\ref{sec:experiments} prove that token-level MIA is useful, powerful and versatile.
For each given sequence $\mathbf{x}=\{x_1x_2\dots x_k\}$, our token-based MIA produces $k-1$ membership scores $\{s_1,\dots, s_{k-1}\}$. To obtain a sequence-based membership score, we need to aggregate them. The simplest way is averaging. Note that the average token-based InfoRMIA scores are not equal to the InfoRMIA score of the entire sequence, as the two operations (InfoRMIA and averaging) are obviously not commutative due to the presence of division. 

A stronger aggregation should depend on the model or underlying data distribution. Such tailored aggregation typically needs to be optimized on additional holdout data. Since the original RMIA also needs to perform a grid search to optimize its hyperparameter $a$ in an offline attack, we can reuse this setting to find the optimal aggregation methods for the given distribution of data and models. However, such a model and dataset specific aggregator that requires additional knowledge and computation power is not always realistic. For practicality, we only evaluate generic aggregation methods, such as averaging and min-$k$, in this paper.
\section{Token-Level Privacy Assessment Interface} \label{sec:viz}
\subsection{Visualizing Information Leakage on the Token Level} \label{sec:tool}
\begin{figure}[!t]
	\centering
	\includegraphics[width=1\linewidth]{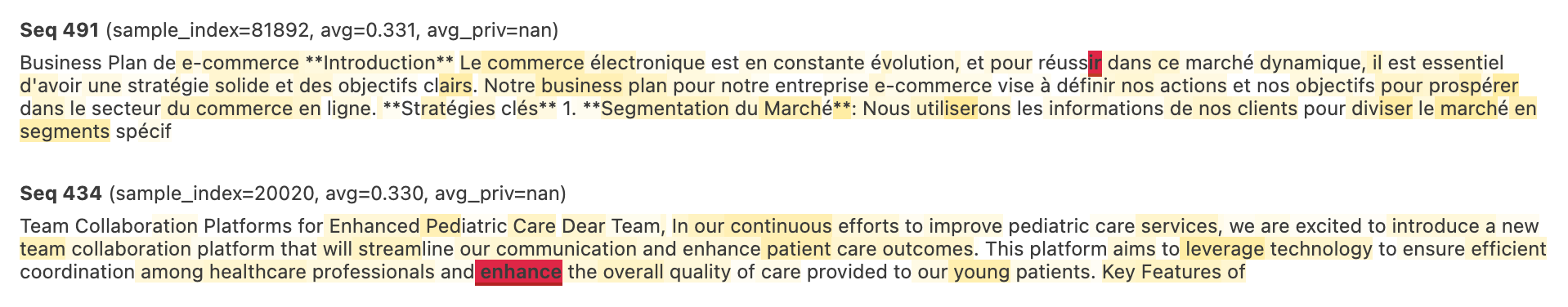}
	\caption{Two of the ten most memorized sequences contain no private tokens. These pose little privacy risk, yet sequence-based frameworks overestimate their risk. See also Figure \ref{fig:top_10_seq}.}
	\label{fig:highseqnoprivate}
\end{figure}

With token-based membership scores, we build a simple HTML interface to visualize a heatmap of token-level memorization over input text (Figures~\ref{fig:viz}, \ref{fig:highseqnoprivate}, \ref{fig:top_10_seq}, \ref{fig:top_10_token}), where the darkness of the highlight reflects the degree of memorization. This fine-grained view enables more accurate privacy assessment, as auditors can directly inspect leakage on the actual private tokens.

We find empirical evidence supporting our intuition that sequence-level signals may not correspond to true privacy risks. Specifically, we observe a low correlation between sequence-level and private-token scores (Figure~\ref{fig:seq_score_vs_token_score}) and discover that many of the most “memorized” sequences either contain disproportionately little private information (Figure~\ref{fig:top_10_seq}) or no private information at all (Figure~\ref{fig:highseqnoprivate}). Conversely, we also find evidence that signals from private tokens are often diluted by the presence of many common tokens in long texts (Figure~\ref{fig:top_10_token}).

This fine-grained analysis is only possible with our token-level framework and highlights the limitations of existing sequence-based notions of privacy. We believe this tool can be highly valuable for practitioners and auditors who need precise, interpretable privacy quantification.

\subsection{Token-Level Analysis Reveals More Than AUCs} \label{sec:pinpoint}
\begin{figure}[!htb]
	\centering
	\includegraphics[width=0.8\linewidth]{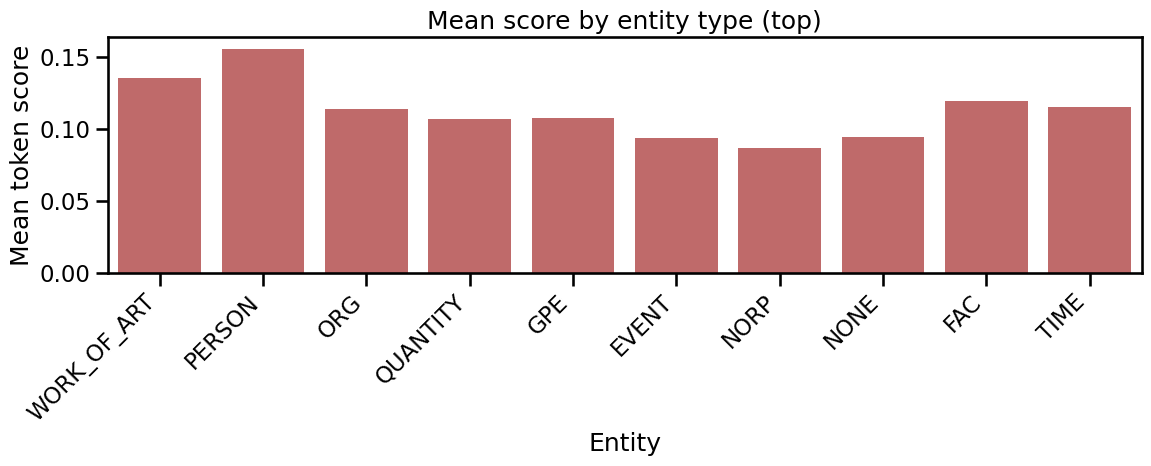}
	\caption{Histogram of the average token scores across the top entity groups on AG News. The ``None'' type represents words that are not nouns.}
	\label{fig:agnews_token_histo}
\end{figure}

Our token-level framework also reveals insights that aggregate metrics like AUC cannot capture. For AG News (Appendix~\ref{app:agnews}), we hypothesize that sensitive information typically appears in personally identifiable information (PII). We therefore use SpaCy~\citep{spacy2} to classify entities. Overall, token-level scores roughly follow a normal distribution (Figure~\ref{fig:agnews_token_dist}). Tokens labeled \verb|PERSON| and \verb|WORK_OF_ART| have the highest average membership scores (Figure~\ref{fig:agnews_token_histo}, Table~\ref{table:agnews_entity_stats}), indicating that names of people and artworks are more likely to be memorized. Examining the top 1\% of the highest-scoring tokens, these two types of tokens also have the highest memorization rate (Table~\ref{table:agnews_entity_stats}), reinforcing that PII is disproportionately more memorized.

\begin{figure}[!ht]
	\centering
	\begin{subfigure}[b]{0.48\textwidth}
		\centering
		\includegraphics[width=0.8\textwidth]{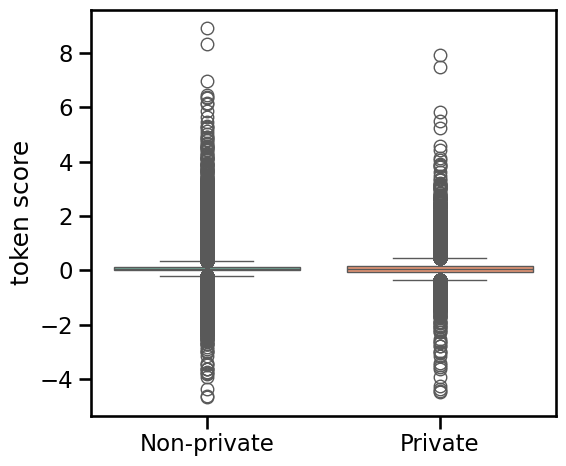}
		\caption{Non-private tokens show a slightly higher mean and larger variance in membership scores compared to private tokens.}
		\label{fig:ai4privacy_boxplot}
	\end{subfigure}
	\hfill
	\begin{subfigure}[b]{0.48\textwidth}
		\includegraphics[width=0.8\linewidth]{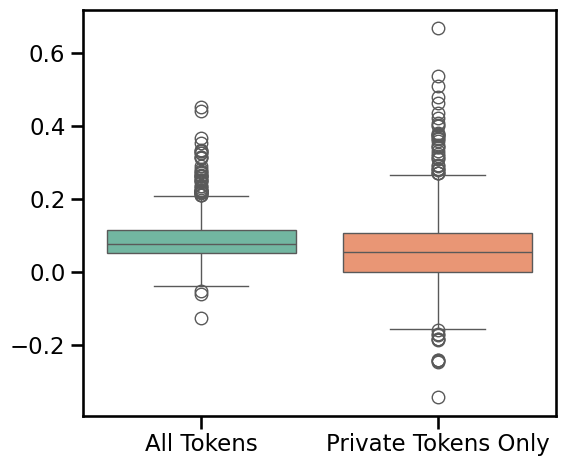}
		\caption{Average scores for all versus private tokens within each sequence. The higher max for private tokens proves their signals get diluted at the sequence level.}
		\label{fig:seq_privatetoken_boxplot}
	\end{subfigure}
	\caption{Boxplots comparing token-level and sequence-level membership scores. More details are provided in Table \ref{table:ai4privacy_stats} and Figure \ref{fig:high_token_ai4privacy_bar}.}
	\label{fig: boxplot}
\end{figure}

For ai4privacy (Appendix~\ref{app:ai4privacy}), each sequence includes a “privacy mask” that marks synthetic personal information. We divide tokens into private and non-private sets and find that the average membership score of private tokens is slightly lower than that of non-private tokens (Table~\ref{table:ai4privacy_stats}, Figure~\ref{fig:ai4privacy_boxplot}). This suggests that the high AUCs reported by sequence-level MIAs may largely reflect memorization of non-private content, which is less relevant for privacy. Hence, AUC alone is a poor indicator of true privacy risk in LLMs.

\section{Experiments} \label{sec:experiments}
\subsection{InfoRMIA Dominates the Original RMIA}
We first show in Table \ref{table:privay_meter_old_new_rmia} that InfoRMIA dominates the original RMIA on all datasets and models across all experiments. For easy benchmarking, we use the new ML Privacy Meter\footnote{\url{https://github.com/privacytrustlab/ml_privacy_meter}}, which is an open-source Python library that audits privacy based on RMIA, released by the same lab behind the RMIA paper\footnote{There are some incorrect implementations of RMIA online. To ensure correctness, we opt for the library from the same lab/authors.}. We evaluate all three default datasets and configurations in the Privacy Meter: Purchase100~\citep{shokri2017membership}, CIFAR-10~\citep{krizhevsky2009learning} and AG News~\citep{zhang2015character}, which cover tabular, image and text datasets. In particular, the AG News dataset is used to train autoregressive GPT-2 models~\citep{radford2019language}, instead of classification models.

\begin{table}[!ht]
\caption{Comparison of AUC and TPR@0.1\%FPR between the original and InfoRMIA, both with 4 reference models on different datasets with different population data sizes. All attacks are offline attacks as the Privacy Meter only implements the offline version for practical considerations. The datasets are loaded in the default way as provided by the Privacy Meter. For CIFAR-10 experiments, the Privacy Meter does not include augmentations in model training and thus attacking, hence the lower numbers compared to the RMIA paper. For Purchase100, the Privacy Meter uses a much larger training set compared to the RMIA paper, which is a better evaluation per ~\citet{suri2024parameters}.}
\label{table:privay_meter_old_new_rmia}
\begin{center}
\resizebox{\textwidth}{!}{%
\begin{tabular}{llllllllll}
\toprule
&& \multicolumn{2}{c}{\bf AG News}  && \multicolumn{2}{c}{\bf CIFAR-10}  && \multicolumn{2}{c}{\bf Purchase100}\\ 
\cmidrule{3-4} \cmidrule{6-7} \cmidrule{9-10}
$|Z|$ && \multicolumn{1}{c}{$100$} & \multicolumn{1}{c}{$1000$} && \multicolumn{1}{c}{$1000$} & \multicolumn{1}{c}{$10000$} && \multicolumn{1}{c}{$1000$} & \multicolumn{1}{c}{$10000$} \\ 
\midrule
\multirow{2}{*}{\bf RMIA}      &AUC            & 0.8574  & 0.8766  && 0.8229 & 0.8327 && 0.5311  & 0.5432   \\ \cmidrule{2-10}
                               &TPR@0.1\%FPR   & 0.00\%  & 1.60\%  && 0.00\% & 0.00\% && 0.00\%  & 0.00\%   
\\ \midrule
\multirow{2}{*}{\bf InfoRMIA} &AUC            & \bf 0.8784  & \bf 0.8784  && \bf 0.8330 & \bf 0.8330 && \bf 0.5754  & \bf 0.5754   \\ \cmidrule{2-10}
                               &TPR@0.1\%FPR   & \bf 12.0\%  & \bf 12.0\%  && \bf 5.82\% & \bf 5.82\% && \bf 0.32\%  & \bf 0.32\%
\\ \bottomrule
\end{tabular}%
}
\end{center}
\end{table}

In addition to higher AUCs, InfoRMIA greatly improves the TPR at very small FPR levels, indicating stronger member identification power without making (many) mistakes. As \citet{carlini2022membership} argued, this metric is a better indication of the membership identification power of an MIA. More importantly, we notice that InfoRMIA is less sensitive to the size of the population data $Z$. This affirms our intuition in the previous section, and is extremely valuable in practice, as the attack can now be run accurately without a large pool of population data or additional resources for computing signals on population data.

\subsection{Solving Sequence-Level Membership Inference with Token-Level Membership Signals}
\paragraph{Finetuned LLMs} We also demonstrate that token-level membership inference can be used to conduct sequence-based membership inference with competitive performance, despite not being designed for it. We show in Table~\ref{table:token-based-rmia} that token-level membership inference with simple averaging yields competitive performance in sequence-level membership inference evaluation. 

\begin{table}[!htb]
\caption{Comparison of AUC and TPR@1\%FPR between the sequence-based and token-based InfoRMIA, and the original RMIA. The description of the datasets used is in the appendix.}
\label{table:token-based-rmia}
\begin{center}
\resizebox{\textwidth}{!}{%
\begin{tabular}{lccccccccc}
\toprule
\bf Datasets                    & \multicolumn{1}{l}{\bf FT Epochs} & \multicolumn{2}{c}{\bf RMIA} && \multicolumn{2}{c}{\bf InfoRMIA} && \multicolumn{2}{c}{\bf InfoRMIA (token)} \\
\cmidrule{3-4} \cmidrule{6-7} \cmidrule{9-10}
                            & \multicolumn{1}{l}{}          & AUC      & TRP@FPR    && AUC         & TRP@FPR      && AUC             & TRP@FPR          \\
\midrule 
AG News                     & 1                             & 0.839    & 0.00\%        && \bf 0.843       & \bf 23.0\%          && 0.836           & 20.2\%              \\
                            & 4                             & \bf 0.945    & 0.00\%        && \bf 0.945       & 16.2\%          && 0.942           & \bf 20.6\%              \\
\midrule 
\multirow{2}{*}{ai4privacy} & 1                             & 0.643    & 6.6\%         && \bf 0.644       & \bf 10.6\%          && 0.620           & 9.0\%               \\
                            & 4                             & 0.821    & 26.0\%        && \bf 0.822       & \bf 27.2\%          && 0.804           & 23.2\%             
\\ \bottomrule
\end{tabular}%
}
\end{center}
\end{table}

\paragraph{Pretrained LLMs} Besides finetuned models, we also evaluate our newly proposed method on pretrained models (Table~\ref{table:mimir-tpr01}-\ref{table:mimir_same_ref}) on the most popular benchmark, MIMIR~\citep{duan2024membership}. As \citet{duan2024membership} pointed out, reference-based MIAs do not perform well on MIMIR due to the lack of quality reference models. Ideally, the reference models should have the same model architecture and be trained on the same data distribution as the target LLM, but with (partially) disjoint datasets (which effectively makes them OUT models in RMIA's terminology). However, for open-source LLMs, it is not possible to find such reference models. Hence, \citet{duan2024membership} concluded that using a model trained on a vast corpus as the reference helps, as it is less fitted to the target data distribution. However, in our experiments, we find that using an earlier snapshot of the LLM is more useful. In particular, we use the \verb|Pythia-160M| checkpoint after the first step as our sole reference model\footnote{We use the non-deduped version as it sees fewer unique training sequences and hence more OUT.}. This is a very practical solution as this checkpoint can be easily trained with lower-end hardware within a short period of time, even if it was not available online. Since no in-distributional population data can be easily obtained, we only evaluate the token-level InfoRMIA in our experiments, which has a similar computational complexity as the \emph{Ref} method~\citep{carlini2021extracting} in MIMIR. The explanation of each method in the table is available in Appendix~\ref{app:mimir}. 

We observe that our token-level InfoRMIA, although not specifically designed for sequence-level membership inference, achieves the strongest membership inference performance~(Table~\ref{table:mimir-tpr01},~\ref{table:mimir-tpr001}) even when using only a single, less ideal reference model. In Tables~\ref{table:mimir-tpr01_same_ref}–\ref{table:mimir_same_ref}, we report results obtained by using the step-1 checkpoint of the target model as the reference and observe minimal change in utility. We further show that our method is the strongest reference-based MIA for pretrained LLMs, outperforming prior reference-model approaches (Ref). We also show that our method is the strongest reference-based MIA for pretrained LLMs, compared to Ref. We also find that simple averaging outperforms the min-$k$ aggregation when targeting high true-positive rates at low false-positive rates (TPR at small FPR). This is expected, as min-$k$ aggregation essentially acts as a non-member detector: non-members tend to contain more low-probability tokens~\citep{shi2024detecting}.Consequently, min-$k$ is less suitable for high precision member detection with minimal error rates. However, when evaluating using AUC (Table~\ref{table:mimir}), the ordering reverses. This aligns with the argument by \citet{carlini2022membership} that AUC can be misleading as a privacy metric. Nonetheless, many recent works evaluating on MIMIR still report only AUC in their main text, which may encourage a suboptimal trend for future development of LLM MIAs.
\section{Conclusion}
In this paper, we propose a new information-theoretic formulation of the membership inference test. The resulting attack, InfoRMIA, consistently outperforms the prior state-of-the-art RMIA across tabular, image, and text datasets, while eliminating RMIA’s reliance on a large population set. Its superior performance stems from a more principled statistical test and the use of a continuous score rather than a discrete one.

We then introduce a new perspective for analyzing privacy risks in LLMs through a token-level analysis framework. It can reveal which tokens are memorized within each sequence, while achieving higher membership inference power. By uncovering fine-grained memorization patterns, our token-level framework enables more precise privacy risk estimation, and opens the door to downstream applications such as targeted machine unlearning and token-guided data reconstruction and extraction. We leave the systematic exploration of these applications to future work.

\bibliography{reference}
\bibliographystyle{plainnat}

\appendix
\section{The Original RMIA} \label{app:rmia}
The original test statistic of RMIA can be written as
\begin{equation} \label{eqn:rmia_original}
	p_z\left(\frac{p(\theta|x)}{p(\theta|z)}\geq \gamma\right),
\end{equation} where $\theta$ is the target model, $x$ is the target query, $z$ is drawn from a population  $Z$ of the same data distribution as the training data, and $\gamma \geq 1$ is a hyperparameter that serves as a threshold.

In simple terms, RMIA counts the proportion of ``similar" data the target $x$ dominates. In practice, the test statistic is written as
\begin{equation}
	p_z\left(\frac{p(x|\theta)}{p(x)}/ \frac{p(z|\theta)}{p(z)}\geq \gamma\right) = \frac{1}{|Z|}\sum_{z \in Z} \mathbb{I}\left(\frac{p(x|\theta)}{p(x)}/ \frac{p(z|\theta)}{p(z)}\geq \gamma\right).
\end{equation}

Note that the formulation in Eqn~\ref{eqn:rmia_original} makes the membership score a discrete value, whose granularity depends on the size of $Z$. Intuitively, the more $z$ data RMIA uses, the finer the ``bins" become, and the more distinguishing and precise the signal gets. Empirically, \citet{zarifzadeh2024low} have also reported this relationship between the size of $Z$ and the attack performance. The empirical insight was that using $Z$ of about $10\%$ of the training set size is sufficient. However, for LLMs, even $10\%$ represents an astronomical number of samples.

\section{Derivation of InfoRMIA Scores}
\begin{equation} \label{eqn:token_info_rmia_deri}
    \begin{aligned}
  &  \sum_{z\in Z} p(z) \log\left(\frac{p(\theta | x)}{p(\theta | z)}\right) \\
  =& \sum_{z\in Z} p(z) \log\left(\frac{p(x|\theta)p(z)}{p(z|\theta)p(x)}\right)  \\ 
  =& \sum_{z\in Z} p(z) \log\left(\frac{p(x|\theta)p(z)}{p(z|\theta)p(x)}\right)+ p(x) \log\left(\frac{p(x|\theta)p(x)}{p(x|\theta)p(x)}\right) \\
  =& \sum_{z\in V} p(z) \log\left(\frac{p(x|\theta)p(z)}{p(z|\theta)p(x)}\right)  \\
 =&  \sum_{z \in V} p(z) \log\left(\frac{p(x|\theta)}{p(x)}\right) + \sum_{z \in V} p(z) \log\left(\frac{p(z)}{p(z|\theta)}\right)\\
 =& \log\left(\frac{p(x|\theta)}{p(x)}\right) + D_{\text{KL}}\left(p(z) \ || \ p(z|\theta)\right)
\end{aligned}
\end{equation}

\section{Implementation Details}
\subsection{RMIA Reference Model Training}
We use the default hyperparameters in ML Privacy Meter to train target and reference models when comparing the original and InfoRMIA, except the number of epochs. For each dataset, the hyperparameter choices can be found in \url{https://github.com/privacytrustlab/ml_privacy_meter/tree/master/configs}. For CIFAR-10 and Purchase-100, we use 100 epochs, while for AG News, we use 1 epoch.

\subsection{Software and Hardware}
For all transformer models and langugage datasets, we use the libraries from Huggingface. All computations are done on two NVIDIA RTX-3090 and two H100 GPUs.
\section{Datasets}
\subsection{AG News} \label{app:agnews}
AG News~\citep{zhang2015character} is a news dataset that contains four categories of news articles. Its training set size is 120,000. In our experiment, we ignore the labels column and train autoregressive models on it.

\subsection{ai4privacy} \label{app:ai4privacy}
The ai4privacy dataset we used is the pii-masking-300k variant, that can be access at \url{https://huggingface.co/datasets/ai4privacy/pii-masking-300k}. It is divided into two parts: OpenPII-220k and FinPII-80k. The FinPII has additional classes that are specific to the Finance and Insurance domains. In this dataset, there is a ``privacy\_mask" column that marks the beginning and end location for each piece of private information. Thus, we can use this information to categorize each token as private or non-private. It also assigns a type to private substrings, such as last names or email addresses, enabling us to do more interesting analysis.

\subsection{The MIMIR Benchmark} \label{app:mimir_dataset}
MIMIR is a benchmark based on the Pile~\citep{pile} dataset, where non-members highly overlapped with any member sequence are removed from the evaluation. Since the members and non-members are randomly shuffled before being split, MIMIR avoids the error of having a large distributional shift between the two sets~\citep{maini2024llm, das2025blind}. It is also one of the most active benchmarks with official implementations of recent methods such as the MinK++~\citep{zhang2025mink}, ReCaLL~\citep{xie2024recall}, and DC-PDD~\citep{zhang2024pretraining}. 

\subsubsection{Explanations of All Attack Methods} \label{app:mimir}
We will briefly explain the score formulation of each attacking method in the MIMIR benchmark. The full details of each attack can be found at the respective papers. Note that in this benchmark, the higher the score is, the less likely it is a member.
\begin{itemize}
    \item \textbf{LOSS}~\citep{yeom2018privacy}: the average loss values of the sequence
    \item \textbf{Zlib}~\citep{carlini2021extracting}: the calibrated loss values by the entropy, estimated by the length after zlib compression
    \item \textbf{Min-K\%}~\citep{shi2024detecting}: the average of the bottom-$k\%$ of token probabilities, and taking the negative (to align the score's sign with MIMIR's standard)
    \item \textbf{Min-K\%++}~\cite{zhang2025mink}: the average of the bottom-$k\%$ of token probabilities calibrated by the mean and variance of each token position's softmax output distributions, and taking the negative
    \item \textbf{DC-PDD}~\citep{zhang2024pretraining}: the average token probabilities calibrated by token frequencies calculated on a reference dataset, and take the negative
    \item \textbf{Ref}~\citep{carlini2021extracting}: the average token loss gap between a reference model and the target model
\end{itemize}

\subsubsection{Why ReCaLL was Excluded in Our Table} \label{app:recall}
The ReCaLL attack pushed to the MIMIR benchmark is a simplied version (according to the author), and very problematic. There is an implicit information leakage about the membership labels in crafting the attack signals, making the result unreliable and unfair. By right, the attacker should not be able to distinguish any non-member from any member, which means the attacker has no information that can be used to tell non-members apart from members before the attack. However, the ReCaLL attack in MIMIR explicitly uses non-members in the evaluation set as its ``non-member" prefix to be prepended to all sequences, making the attacker aware of the membership label of certain non-members. Although the label information is not explicitly used in the attack, it is an implicit information leak that can be used to tell apart the two sets. Hence, the current implementation of the ReCaLL attack in MIMIR is not correct. We will include it once the official implementation is fixed.
\section{Additional Results}
\subsection{MIMIR results}
We provide more results on the MIMIR benchmark here. In particular, we use the \verb|ngram<0.8| split. Table~\ref{table:mimir-tpr01} and \ref{table:mimir-tpr001} show the results on TPR @1\% FPR and TPR@0.1\%FPR respectively, while Table~\ref{table:mimir} shows the AUCs. The results show that our method has strongest inference power (highest TPR at small FPRs), while achieving very competitive results on AUCs. It is also stronger than the prior reference-based MIA, Ref~\citep{carlini2021extracting}.

\begin{table}[ht!]
	\caption{TPR @1\% FPR on MIMIR with deduped Pythia models. on MIMIR benchmark with deduped Pythia models. The \emph{Neighbor} method is not included due to its computational complexity and relatively inferior performance reported in prior works. \emph{ReCaLL} is not included for reasons in Appendix~\ref{app:recall}. \emph{Ref} method is evaluated using the checkpoint of Pythia-160m after the first step as the reference model. Our method (\emph{InfoRMIA}) is the token-based InfoRMIA that does not require additional population data. InfoRMIA1 uses averaging to aggregate, while InfoRMIA2 uses min-k\%, using the same hyperparameter $k$ as \emph{Min-K\%} and \emph{Min-K\%++}. Bold numbers are the best, and the underlined are the best reference-based.}
	\label{table:mimir-tpr01}
	\begin{center}
		\scriptsize
		\setlength{\tabcolsep}{2pt}
		\renewcommand{\arraystretch}{1.15}
		
		\resizebox{\textwidth}{!}{%
			\begin{tabular}{l *{16}{c}}
				\toprule
				& \multicolumn{4}{c}{\textbf{Wikipedia}}
				& \multicolumn{4}{c}{\textbf{Github}}
				& \multicolumn{4}{c}{\textbf{Pile CC}}
				& \multicolumn{4}{c}{\textbf{PubMed Central}} \\
				\cmidrule(lr){2-5}\cmidrule(lr){6-9}\cmidrule(lr){10-13}\cmidrule(lr){14-17}
				\textbf{Method}
				& 160M & 1.4B & 2.8B & 6.9B
				& 160M & 1.4B & 2.8B & 6.9B
				& 160M & 1.4B & 2.8B & 6.9B
				& 160M & 1.4B & 2.8B & 6.9B \\
				\midrule
				Loss         & 0.9 & 0.6 & 0.6 & 0.6 & 13.1 & 13.3 & 21.9 & 13.2 & 0.4 & 0.7 & 0.8 & 0.9 & 0.7 & 0.4 & 0.6 & 0.4 \\
				Zlib         & 1.3 & 0.7 & 0.8 & 0.6 & 14.3 & 16.9 & \textbf{24.0} & \textbf{15.5} & 0.7 & 0.7 & 0.9 & \textbf{1.5} & 0.3 & 0.4 & 0.5 & 0.4 \\
				Min-K\%      & \textbf{1.4} & 0.9 & 0.6 & 0.5 & 12.0 & 13.1 & 21.8 & 13.0 & 0.5 & 0.6 & 0.7 & 1.0 & 0.6 & 0.2 & 0.6 & 0.4 \\
				Min-K\%++    & 1.2 & 0.7 & 0.6 & 1.0 & 11.2 & 12.8 & 18.1 & 12.8 & 1.1 & \textbf{1.1} & \textbf{1.2} & \textbf{1.5} & 0.6 & 0.4 & 0.5 & 0.6 \\
				DC-PDD       & 0.9 & 0.4 & \textbf{1.2} & \textbf{1.4} & 10.8 & 11.3 & 9.8 & 10.7 & 0.4 & \textbf{1.1} & 0.6 & 1.1 & \textbf{1.5} & 0.8 & \textbf{1.3} & 1.3 \\
				Ref          & 0.9 & 0.8 & 0.7 & 0.6 & 13.4 & 13.9 & 20.3 & 14.7 & 0.6 & 0.7 & \underline{0.8} & \underline{1.0} & 0.8 & 0.6 & 0.5 & 0.4 \\
				\rowcolor{RowGray}
				InfoRMIA1   & 0.8 & 0.9 & 0.6 & \underline{0.9} & \underline{\textbf{14.7}} & \underline{\textbf{17.7}} & \underline{21.2} & \underline{\textbf{15.5}} & \underline{\textbf{1.2}} & \underline{0.9} & 0.7 & 0.6 & 1.0 & 0.5 & 0.3 & 0.4 \\
				\rowcolor{RowGray}
				InfoRMIA2   & \underline{1.1} & \underline{\textbf{1.2}} & \underline{0.9} & 0.7 & 13.0 & 14.1 & 18.3 & 14.5 & 0.4 & 0.5 & 0.7 & 0.5 & \underline{1.2} & \underline{\textbf{1.4}} & \underline{\textbf{1.3}} & \underline{\textbf{1.6}} \\
				
				\toprule
				& \multicolumn{4}{c}{\textbf{ArXiv}}
				& \multicolumn{4}{c}{\textbf{DM Mathematics}}
				& \multicolumn{4}{c}{\textbf{HackerNews}}
				& \multicolumn{4}{c}{\textbf{Average}} \\
				\cmidrule(lr){2-5}\cmidrule(lr){6-9}\cmidrule(lr){10-13}\cmidrule(lr){14-17}
				\textbf{Method}
				& 160M & 1.4B & 2.8B & 6.9B
				& 160M & 1.4B & 2.8B & 6.9B
				& 160M & 1.4B & 2.8B & 6.9B
				& 160M & 1.4B & 2.8B & 6.9B \\
				\midrule
				
				Loss         & 0.7 & 0.7 & 0.4 & 0.8 & 0.5 & 0.5 & 1.1 & \textbf{1.1} & 0.9 & 0.7 & 0.6 & 0.8 & 2.5 & 2.4 & 3.7 & 2.5 \\
				Zlib         & 0.5 & 0.2 & 0.4 & 0.7 & 1.1 & 0.9 & 0.9 & 0.6 & 0.6 & 1.0 & 1.0 & 1.0 & 2.7 & 3.0 & \textbf{4.1} & \underline{\textbf{2.9}} \\
				Min-K\%      & 0.3 & 0.3 & 0.4 & 0.7 & 0.8 & 0.6 & 0.2 & 0.4 & 0.7 & 0.9 & 0.7 & 1.1 & 2.3 & 2.4 & 3.6 & 2.4 \\
				Min-K\%++    & \textbf{1.1} & \textbf{1.9} & \textbf{1.2} & \textbf{1.4} & 1.0 & 1.0 & \textbf{1.2} & 1.0 & 0.7 & 0.5 & 1.1 & 0.7 & 2.4 & 2.6 & 3.4 & 2.7 \\
				DC-PDD       & 0.5 & 1.0 & 0.9 & 0.5 & 0.5 & 0.4 & 0.2 & 0.1 & 1.3 & 1.0 & 0.5 & 1.2 & 2.3 & 2.3 & 2.1 & 2.3 \\
				Ref          & \underline{0.8} & \underline{0.5} & \underline{0.5} & \underline{0.6} & \underline{\textbf{1.2}} & 1.0 & \underline{\textbf{1.2}} & 0.7 & 1.4 & 0.7 & 0.7 & 0.7 & 2.7 & 2.6 & 3.5 & 2.7 \\
				\rowcolor{RowGray}
				InfoRMIA1   & 0.3 & \underline{0.5} & 0.4 & \underline{0.6} & 0.7 & \underline{\textbf{1.3}} & 1.0 & \underline{\textbf{1.1}} & 1.5 & \underline{\textbf{1.2}} & \underline{\textbf{1.7}} & 1.3 & \underline{\textbf{2.9}} & \underline{\textbf{3.3}} & \underline{3.7} & \underline{\textbf{2.9}} \\
				\rowcolor{RowGray}
				InfoRMIA2   & 0.1 & 0.3 & 0.3 & 0.4 & \underline{\textbf{1.2}} & 0.8 & 1.1 & 0.7 & \underline{\textbf{1.7}} & \underline{\textbf{1.2}} & 1.0 & \underline{\textbf{1.8}} & 2.7 & 2.8 & 3.4 & \underline{\textbf{2.9}} \\
				\bottomrule
			\end{tabular}%
		}
	\end{center}
\end{table}

\begin{table}[ht!]
	\caption{TPR @0.1\% FPR on MIMIR with deduped Pythia models.}
	\label{table:mimir-tpr001}
	\begin{center}
		\scriptsize
		\setlength{\tabcolsep}{2pt}
		\renewcommand{\arraystretch}{1.15}
		
		\resizebox{\textwidth}{!}{%
			\begin{tabular}{l *{16}{c}}
				\toprule
				& \multicolumn{4}{c}{\textbf{Wikipedia}}
				& \multicolumn{4}{c}{\textbf{Github}}
				& \multicolumn{4}{c}{\textbf{Pile CC}}
				& \multicolumn{4}{c}{\textbf{PubMed Central}} \\
				\cmidrule(lr){2-5}\cmidrule(lr){6-9}\cmidrule(lr){10-13}\cmidrule(lr){14-17}
				\textbf{Method}
				& 160M & 1.4B & 2.8B & 6.9B
				& 160M & 1.4B & 2.8B & 6.9B
				& 160M & 1.4B & 2.8B & 6.9B
				& 160M & 1.4B & 2.8B & 6.9B \\
				\midrule
				Loss         & 0.0 & 0.1 & \textbf{0.2} & 0.1 & 5.7 & 4.8 & 7.9 & 5.2 & 0.0 & \textbf{0.2} & 0.1 & 0.2 & \textbf{0.0} & 0.0 & 0.0 & 0.0 \\
				Zlib         & \textbf{0.1} & 0.1 & 0.1 & 0.1 & \textbf{8.4} & \textbf{5.7} & \textbf{8.6} & \textbf{6.9} & 0.0 & \textbf{0.2} & 0.2 & \textbf{0.3} & \textbf{0.0} & 0.0 & 0.0 & 0.0 \\
				Min-K\%      & 0.0 & 0.1 & \textbf{0.2} & 0.1 & 5.7 & 4.8 & 8.0 & 4.9 & 0.0 & \textbf{0.2} & 0.1 & 0.2 & \textbf{0.0} & 0.0 & 0.0 & 0.0 \\
				Min-K\%++    & 0.0 & 0.0 & 0.1 & 0.0 & 6.1 & 3.5 & 4.6 & 2.1 & \textbf{0.1} & \textbf{0.2} & 0.1 & \textbf{0.3} & \textbf{0.0} & 0.0 & 0.0 & 0.0 \\
				DC-PDD       & 0.0 & 0.0 & 0.0 & 0.0 & 3.7 & 0.3 & 0.3 & 1.1 & 0.0 & 0.0 & \textbf{0.3} & 0.2 & \textbf{0.0} & 0.0 & 0.1 & 0.1 \\
				Ref          & 0.0 & 0.0 & 0.0 & 0.0 & 4.3 & \underline{4.3} & \underline{2.2} & \underline{3.5} & \underline{\textbf{0.1}} & \underline{\textbf{0.2}} & \underline{\textbf{0.3}} & \underline{\textbf{0.3}} & \underline{\textbf{0.0}} & \underline{\textbf{0.1}} & 0.0 & 0.0 \\
				\rowcolor{RowGray}
				InfoRMIA1   & \underline{\textbf{0.1}} & \underline{\textbf{0.2}} & \underline{\textbf{0.2}} & \underline{\textbf{0.2}} & 0.0 & 0.0 & 0.6 & 1.0 & \underline{\textbf{0.1}} & 0.1 & 0.1 & 0.1 & \underline{\textbf{0.0}} & \underline{\textbf{0.1}} & \underline{\textbf{0.3}} & \underline{\textbf{0.3}} \\
				\rowcolor{RowGray}
				InfoRMIA2   & 0.0 & 0.0 & 0.1 & 0.1 & \underline{4.8} & 0.0 & 0.9 & 0.2 & \underline{\textbf{0.1}} & 0.1 & 0.1 & 0.2 & \underline{\textbf{0.0}} & 0.0 & 0.0 & 0.0 \\
				
				\toprule
				& \multicolumn{4}{c}{\textbf{ArXiv}}
				& \multicolumn{4}{c}{\textbf{DM Mathematics}}
				& \multicolumn{4}{c}{\textbf{HackerNews}}
				& \multicolumn{4}{c}{\textbf{Average}} \\
				\cmidrule(lr){2-5}\cmidrule(lr){6-9}\cmidrule(lr){10-13}\cmidrule(lr){14-17}
				\textbf{Method}
				& 160M & 1.4B & 2.8B & 6.9B
				& 160M & 1.4B & 2.8B & 6.9B
				& 160M & 1.4B & 2.8B & 6.9B
				& 160M & 1.4B & 2.8B & 6.9B \\
				\midrule
				
				Loss         & 0.1 & 0.0 & 0.0 & \textbf{0.1} & 0.0 & 0.0 & 0.2 & 0.0 & 0.1 & 0.0 & 0.0 & 0.0 & 0.8 & 0.7 & 1.2 & 0.8 \\
				Zlib         & 0.0 & 0.0 & 0.0 & 0.0 & 0.0 & 0.0 & 0.0 & 0.0 & 0.1 & 0.2 & 0.2 & 0.2 & \textbf{1.2} & \textbf{0.9} & \textbf{1.3} & \textbf{1.1} \\
				Min-K\%      & 0.0 & 0.0 & 0.0 & 0.0 & 0.0 & 0.0 & 0.0 & 0.0 & \textbf{0.2} & 0.1 & 0.1 & 0.1 & 0.8 & 0.7 & 1.2 & 0.8 \\
				Min-K\%++    & 0.0 & 0.0 & 0.0 & 0.0 & 0.1 & 0.0 & \textbf{0.5} & 0.2 & \textbf{0.2} & 0.0 & 0.1 & 0.0 & 0.9 & 0.5 & 0.8 & 0.4 \\
				DC-PDD       & 0.0 & 0.0 & \textbf{0.2} & 0.0 & 0.0 & 0.0 & 0.0 & 0.0 & 0.0 & 0.2 & 0.1 & 0.0 & 0.5 & 0.1 & 0.1 & 0.2 \\
				Ref          & \underline{\textbf{0.2}} & \underline{\textbf{0.1}} & \underline{0.0} & \underline{0.0} & \underline{\textbf{0.2}} & 0.0 & 0.1 & 0.1 & \underline{0.1} & 0.0 & 0.0 & 0.2 & 0.7 & \underline{0.7} & \underline{0.4} & \underline{0.6} \\
				\rowcolor{RowGray}
				InfoRMIA1   & 0.1 & 0.0 & \underline{0.0} & \underline{0.0} & 0.1 & \underline{\textbf{0.3}} & \underline{0.3} & \underline{\textbf{0.6}} & 0.0 & 0.0 & 0.0 & 0.0 & 0.1 & 0.1 & 0.2 & 0.3 \\
				\rowcolor{RowGray}
				InfoRMIA2   & 0.0 & 0.0 & \underline{0.0} & \underline{0.0} & 0.0 & 0.0 & 0.0 & 0.0 & \underline{0.1} & \underline{\textbf{0.3}} & \underline{\textbf{0.3}} & \underline{\textbf{0.3}} & \underline{0.7} & 0.1 & 0.2 & 0.1 \\
				\bottomrule
			\end{tabular}%
		}
	\end{center}
\end{table}

\begin{table}[ht!]
\caption{AUC results on MIMIR benchmark with deduped Pythia models.}
\label{table:mimir}
\begin{center}
\scriptsize
\setlength{\tabcolsep}{2pt}
\renewcommand{\arraystretch}{1.15}

\resizebox{\textwidth}{!}{%
\begin{tabular}{l *{16}{c}}
\toprule
& \multicolumn{4}{c}{\textbf{Wikipedia}}
& \multicolumn{4}{c}{\textbf{Github}}
& \multicolumn{4}{c}{\textbf{Pile CC}}
& \multicolumn{4}{c}{\textbf{PubMed Central}} \\
\cmidrule(lr){2-5}\cmidrule(lr){6-9}\cmidrule(lr){10-13}\cmidrule(lr){14-17}
\textbf{Method}
& 160M & 1.4B & 2.8B & 6.9B
& 160M & 1.4B & 2.8B & 6.9B
& 160M & 1.4B & 2.8B & 6.9B
& 160M & 1.4B & 2.8B & 6.9B \\
\midrule
Loss         & 50.2 & 51.0 & 51.7 & 51.6 & 63.7 & 65.8 & 71.2 & 67.6 & 49.5 & 50.1 & 50.1 & 51.3 & 49.9 & 49.8 & 49.9 & 50.5 \\
Zlib         & \textbf{51.0} & 51.8 & 52.4 & 52.3 & \textbf{65.6} & \textbf{67.2} & \textbf{72.2} & 68.8 & 49.6 & 50.2 & 50.3 & 51.2 & 50.0 & 50.0 & 50.0 & 50.6 \\
Min-K\%      & 48.6 & 50.6 & 51.6 & 51.4 & 63.6 & 65.9 & 71.4 & 68.0 & 50.0 & 51.0 & 50.5 & 51.9 & 50.4 & 50.2 & 50.4 & 51.0 \\
Min-K\%++    & 47.7 & \textbf{52.3} & \textbf{53.7} & \textbf{52.4} & 61.4 & 65.7 & 70.7 & \textbf{69.1} & 49.8 & 51.1 & 49.9 & 51.7 & 50.9 & 50.6 & \textbf{51.2} & \textbf{52.3} \\
DC-PDD       & 49.0 & 50.6 & 52.4 & 51.8 & 64.9 & 66.2 & 71.4 & 69.0 & 49.6 & 51.1 & \textbf{51.2} & \textbf{51.9} & 50.5 & \textbf{51.0} & 50.6 & 51.1 \\
Ref          & 50.0 & 50.8 & \underline{51.6} & \underline{51.4} & 63.9 & 66.0 & \underline{71.4} & \underline{67.9} & 49.4 & 50.0 & 50.0 & 51.2 & 49.8 & 49.7 & 49.8 & 50.4 \\
\rowcolor{RowGray}
InfoRMIA1   & \underline{50.9} & \underline{50.8} & 51.0 & 51.2 & \underline{65.0} & \underline{66.1} & 70.8 & 67.0 & 49.4 & 49.6 & 49.8 & 50.5 & 50.2 & 49.7 & 49.5 & 49.8 \\
\rowcolor{RowGray}
InfoRMIA2   & 50.0 & 50.3 & 51.1 & 51.1 & 63.5 & 65.3 & 70.6 & 66.9 & \underline{\textbf{50.6}} & \underline{\textbf{51.1}} & \underline{50.8} & \underline{51.7} & \underline{\textbf{51.4}} & \underline{50.4} & \underline{50.2} & \underline{50.7} \\

\toprule
& \multicolumn{4}{c}{\textbf{ArXiv}}
& \multicolumn{4}{c}{\textbf{DM Mathematics}}
& \multicolumn{4}{c}{\textbf{HackerNews}}
& \multicolumn{4}{c}{\textbf{Average}} \\
\cmidrule(lr){2-5}\cmidrule(lr){6-9}\cmidrule(lr){10-13}\cmidrule(lr){14-17}
\textbf{Method}
& 160M & 1.4B & 2.8B & 6.9B
& 160M & 1.4B & 2.8B & 6.9B
& 160M & 1.4B & 2.8B & 6.9B
& 160M & 1.4B & 2.8B & 6.9B \\
\midrule

Loss         & 50.7 & 51.4 & 51.9 & 52.5 & 49.0 & 48.6 & 48.3 & 48.4 & 49.2 & 50.4 & 51.2 & 51.7 & 51.8 & 52.4 & 53.5 & 53.4 \\
Zlib         & 50.0 & 50.8 & 51.3 & 51.8 & 48.2 & 48.1 & 48.0 & 48.1 & 49.6 & 50.2 & 50.9 & 51.0 & 52.0 & 52.6 & 53.6 & 53.4 \\
Min-K\%      & 50.0 & 51.2 & 52.2 & 52.7 & 49.4 & 49.3 & 49.1 & 49.3 & 50.2 & 51.3 & 52.4 & 53.0 & 51.7 & 52.8 & 53.9 & 53.9 \\
Min-K\%++    & 48.7 & 51.2 & \textbf{53.1} & 52.8 & \textbf{49.9} & \textbf{50.0} & \textbf{50.3} & \textbf{50.2} & \textbf{50.9} & 51.1 & 52.3 & 53.7 & 51.3 & \textbf{53.1} & 54.4 & \textbf{54.6} \\
DC-PDD       & 50.4 & \textbf{52.0} & 52.9 & \textbf{52.9} & 49.0 & 49.3 & 49.8 & 49.7 & 50.7 & 51.8 & \textbf{53.0} & \textbf{53.9} & 52.0 & 53.1 & \textbf{54.5} & 54.3 \\
Ref          & 50.3 & 51.0 & 51.5 & 52.1 & 48.8 & 48.5 & 48.3 & 48.3 & 49.1 & 50.4 & 51.2 & 51.7 & 51.6 & 52.3 & 53.4 & 53.3 \\
\rowcolor{RowGray}
InfoRMIA1   & 50.3 & 51.1 & 51.1 & 51.5 & 48.0 & 47.6 & 47.9 & 47.9 & 50.4 & 50.7 & 51.0 & 51.3 & 52.0 & 52.2 & 53.0 & 52.7 \\
\rowcolor{RowGray}
InfoRMIA2   & \underline{\textbf{50.8}} & \underline{51.2} & \underline{51.6} & \underline{52.3} & \underline{49.0} & \underline{49.1} & \underline{49.0} & \underline{48.9} & \underline{50.4} & \underline{\textbf{51.9}} & \underline{52.4} & \underline{53.1} & \underline{\textbf{52.2}} & \underline{52.7} & \underline{53.7} & \underline{53.5} \\
\bottomrule
\end{tabular}%
}
\end{center}
\end{table}

\subsection{MIMIR results when using the first step checkpoint of the target model as the reference}
Instead of using the first step checkpoint of \verb|Pythia-160m|, we use the checkpoint corresponding to the target model, e.g., we use the \verb|Pythia-1.4b:step1| as the reference when attacking \verb|Pythia-1.4b-deduped|. We found in Table~\ref{table:mimir-tpr01_same_ref}, \ref{table:mimir-tpr001_same_ref} and \ref{table:mimir_same_ref} that the difference in utility is minimal. Therefore, it might be better to stick to snapshots of smaller models as reference models for cost-sensitive auditors. Moreover, using the first step checkpoint of the small LLM as the reference has another benefit: if the checkpoint is not publicly available, training the small LLM for one step is computationally cheap. Normal users can obtain the checkpoint and run the attack with low end computes and short training period.

\begin{table}[ht!]
	\caption{TPR @1\% FPR on MIMIR benchmark with deduped Pythia models when using the first step checkpoint of the target model as the reference.}
	\label{table:mimir-tpr01_same_ref}
	\begin{center}
		\scriptsize
		\setlength{\tabcolsep}{2pt}
		\renewcommand{\arraystretch}{1.15}
		
		\resizebox{\textwidth}{!}{%
			\begin{tabular}{l *{16}{c}}
				\toprule
				& \multicolumn{4}{c}{\textbf{Wikipedia}} & \multicolumn{4}{c}{\textbf{Github}} & \multicolumn{4}{c}{\textbf{Pile CC}} & \multicolumn{4}{c}{\textbf{PubMed Central}} \\
				\cmidrule(lr){2-5} \cmidrule(lr){6-9} \cmidrule(lr){10-13} \cmidrule(lr){14-17}
				\textbf{Method}& 160M & 1.4B & 2.8B & 6.9B & 160M & 1.4B & 2.8B & 6.9B & 160M & 1.4B & 2.8B & 6.9B & 160M & 1.4B & 2.8B & 6.9B \\
				\midrule
				Loss         & 0.9 & 0.6 & 0.6 & 0.6 & 13.1 & 13.3 & 21.9 & 13.2 & 0.4 & 0.7 & 0.8 & 0.9 & 0.7 & 0.4 & 0.6 & 0.4 \\
				Zlib         & 1.3 & 0.7 & 0.8 & 0.6 & 14.3 & 16.9 & \textbf{24.0} & \textbf{15.5} & 0.7 & 0.7 & 0.9 & \textbf{1.5} & 0.3 & 0.4 & 0.5 & 0.4 \\
				Min-K\%      & \textbf{1.4} & 0.9 & 0.6 & 0.5 & 12.0 & 13.1 & 21.8 & 13.0 & 0.5 & 0.6 & 0.7 & 1.0 & 0.6 & 0.2 & 0.6 & 0.4 \\
				Min-K\%++    & 1.2 & 0.7 & 0.6 & 1.0 & 11.2 & 12.8 & 18.1 & 12.8 & 1.1 & \textbf{1.1} & \textbf{1.2} & \textbf{1.5} & 0.6 & 0.4 & 0.5 & 0.6 \\
				DC-PDD       & 0.9 & 0.4 & \textbf{1.2} & \textbf{1.4} & 10.8 & 11.3 & 9.8 & 10.7 & 0.4 & \textbf{1.1} & 0.6 & 1.1 & \textbf{1.5} & 0.8 & \textbf{1.3} & \textbf{1.3} \\
				Ref          & 0.9 & 0.8 & 0.7 & 0.9 & 13.4 & 10.9 & 17.9 & 4.7 & 0.6 & 0.5 & \underline{0.7} & \underline{1.2} & 0.8 & 0.9 & 0.2 & 0.5 \\
				\rowcolor{RowGray}
				InfoRMIA1   & 0.8 & 1.0 & 0.7 & \underline{1.0} & \underline{\textbf{14.7}} & \underline{\textbf{17.5}} & \underline{21.0} & \underline{14.9} & \underline{\textbf{1.2}} & \underline{0.9} & \underline{0.7} & 0.7 & 1.0 & 0.7 & 0.5 & 0.7 \\
				\rowcolor{RowGray}
				InfoRMIA2   & \underline{1.1} & \underline{\textbf{1.3}} & \underline{0.9} & \underline{1.0} & 13.0 & 13.8 & 18.7 & 14.5 & 0.4 & 0.5 & 0.5 & 0.6 & \underline{1.2} & \underline{\textbf{1.2}} & \underline{\textbf{1.3}} & \underline{0.8} \\
				
				\toprule
				& \multicolumn{4}{c}{\textbf{ArXiv}} & \multicolumn{4}{c}{\textbf{DM Mathematics}} & \multicolumn{4}{c}{\textbf{HackerNews}} & \multicolumn{4}{c}{\textbf{Average}} \\
				\cmidrule(lr){2-5} \cmidrule(lr){6-9} \cmidrule(lr){10-13} \cmidrule(lr){14-17}
				\textbf{Method}& 160M & 1.4B & 2.8B & 6.9B & 160M & 1.4B & 2.8B & 6.9B & 160M & 1.4B & 2.8B & 6.9B & 160M & 1.4B & 2.8B & 6.9B \\
				\midrule
				
				Loss         & 0.7 & 0.7 & 0.4 & 0.8 & 0.5 & 0.5 & 1.1 & 1.1 & 0.9 & 0.7 & 0.6 & 0.8 & 2.5 & 2.4 & 3.7 & 2.5 \\
				Zlib         & 0.5 & 0.2 & 0.4 & 0.7 & 1.1 & 0.9 & 0.9 & 0.6 & 0.6 & 1.0 & 1.0 & 1.0 & 2.7 & 3.0 & \textbf{4.1} & \textbf{2.9} \\
				Min-K\%      & 0.3 & 0.3 & 0.4 & 0.7 & 0.8 & 0.6 & 0.2 & 0.4 & 0.7 & 0.9 & 0.7 & 1.1 & 2.3 & 2.4 & 3.6 & 2.4 \\
				Min-K\%++    & \textbf{1.1} & \textbf{1.9} & \textbf{1.2} & \textbf{1.4} & 1.0 & 1.0 & \textbf{1.2} & 1.0 & 0.7 & 0.5 & 1.1 & 0.7 & 2.4 & 2.6 & 3.4 & 2.7 \\
				DC-PDD       & 0.5 & 1.0 & 0.9 & 0.5 & 0.5 & 0.4 & 0.2 & 0.1 & 1.3 & 1.0 & 0.5 & 1.2 & 2.3 & 2.3 & 2.1 & 2.3 \\
				Ref          & \underline{0.8} & 0.3 & \underline{0.6} & \underline{0.5} & \underline{\textbf{1.2}} & 0.9 & 1.0 & 0.3 & 1.4 & 1.0 & 0.8 & 1.0 & 2.7 & 2.2 & 3.1 & 1.3 \\
				\rowcolor{RowGray}
				InfoRMIA1   & 0.3 & \underline{0.4} & 0.3 & 0.4 & 0.7 & \underline{\textbf{1.4}} & 1.1 & \underline{\textbf{1.2}} & 1.5 & \underline{\textbf{1.8}} & \underline{\textbf{1.2}} & 1.3 & \underline{\textbf{2.9}} & \underline{\textbf{3.4}} & \underline{3.6} & \underline{\textbf{2.9}} \\
				\rowcolor{RowGray}
				InfoRMIA2   & 0.1 & 0.3 & 0.3 & 0.3 & \underline{\textbf{1.2}} & 0.9 & \underline{\textbf{1.2}} & 1.0 & \underline{\textbf{1.7}} & 1.3 & 1.0 & \underline{\textbf{1.7}} & 2.7 & 2.8 & 3.4 & 2.8 \\
				\bottomrule
			\end{tabular}%
		}
	\end{center}
\end{table}

\begin{table}[ht!]
	\caption{TPR @0.1\% FPR on MIMIR benchmark with deduped Pythia models when using the first step checkpoint of the target model as the reference.}
	\label{table:mimir-tpr001_same_ref}
	\begin{center}
		\scriptsize
		\setlength{\tabcolsep}{2pt}
		\renewcommand{\arraystretch}{1.15}
		
		\resizebox{\textwidth}{!}{%
			\begin{tabular}{l *{16}{c}}
				\toprule
				& \multicolumn{4}{c}{\textbf{Wikipedia}} & \multicolumn{4}{c}{\textbf{Github}} & \multicolumn{4}{c}{\textbf{Pile CC}} & \multicolumn{4}{c}{\textbf{PubMed Central}} \\
				\cmidrule(lr){2-5} \cmidrule(lr){6-9} \cmidrule(lr){10-13} \cmidrule(lr){14-17}
				\textbf{Method}& 160M & 1.4B & 2.8B & 6.9B & 160M & 1.4B & 2.8B & 6.9B & 160M & 1.4B & 2.8B & 6.9B & 160M & 1.4B & 2.8B & 6.9B \\
				\midrule
				Loss         & 0.0 & \textbf{0.1} & \textbf{0.2} & 0.1 & 5.7 & 4.8 & 7.9 & 5.2 & 0.0 & \textbf{0.2} & 0.1 & 0.2 & \textbf{0.0} & 0.0 & 0.0 & 0.0 \\
				Zlib         & \textbf{0.1} & \textbf{0.1} & 0.1 & 0.1 & \textbf{8.4} & \textbf{5.7} & \textbf{8.6} & \textbf{6.9} & 0.0 & \textbf{0.2} & 0.2 & \textbf{0.3} & \textbf{0.0} & 0.0 & 0.0 & 0.0 \\
				Min-K\%      & 0.0 & \textbf{0.1} & \textbf{0.2} & 0.1 & 5.7 & 4.8 & 8.0 & 4.9 & 0.0 & \textbf{0.2} & 0.1 & 0.2 & \textbf{0.0} & 0.0 & 0.0 & 0.0 \\
				Min-K\%++    & 0.0 & 0.0 & 0.1 & 0.0 & 6.1 & 3.5 & 4.6 & 2.1 & \textbf{0.1} & \textbf{0.2} & 0.1 & \textbf{0.3} & \textbf{0.0} & 0.0 & 0.0 & 0.0 \\
				DC-PDD       & 0.0 & 0.0 & 0.0 & 0.0 & 3.7 & 0.3 & 0.3 & 1.1 & 0.0 & 0.0 & \textbf{0.3} & 0.2 & \textbf{0.0} & 0.0 & 0.1 & 0.1 \\
				Ref          & 0.0 & \underline{\textbf{0.1}} & 0.0 & 0.1 & 4.3 & \underline{0.0} & 0.6 & 0.1 & \underline{\textbf{0.1}} & \underline{0.1} & \underline{0.1} & \underline{0.2} & \underline{\textbf{0.0}} & 0.0 & 0.0 & 0.0 \\
				\rowcolor{RowGray}
				InfoRMIA1   & \underline{\textbf{0.1}} & \underline{\textbf{0.1}} & \underline{\textbf{0.2}} & \underline{\textbf{0.3}} & 0.0 & \underline{0.0} & 0.4 & 0.1 & \underline{\textbf{0.1}} & \underline{0.1} & \underline{0.1} & 0.1 & \underline{\textbf{0.0}} & \underline{\textbf{0.1}} & \underline{\textbf{0.3}} & \underline{\textbf{0.3}} \\
				\rowcolor{RowGray}
				InfoRMIA2   & 0.0 & 0.0 & 0.1 & 0.0 & \underline{4.8} & \underline{0.0} & \underline{0.8} & \underline{0.2} & \underline{\textbf{0.1}} & \underline{0.1} & \underline{0.1} & 0.1 & \underline{\textbf{0.0}} & 0.0 & 0.0 & 0.0 \\
				
				\toprule
				& \multicolumn{4}{c}{\textbf{ArXiv}} & \multicolumn{4}{c}{\textbf{DM Mathematics}} & \multicolumn{4}{c}{\textbf{HackerNews}} & \multicolumn{4}{c}{\textbf{Average}} \\
				\cmidrule(lr){2-5} \cmidrule(lr){6-9} \cmidrule(lr){10-13} \cmidrule(lr){14-17}
				\textbf{Method}& 160M & 1.4B & 2.8B & 6.9B & 160M & 1.4B & 2.8B & 6.9B & 160M & 1.4B & 2.8B & 6.9B & 160M & 1.4B & 2.8B & 6.9B \\
				\midrule
				
				Loss         & 0.1 & 0.0 & 0.0 & \textbf{0.1} & 0.0 & 0.0 & 0.2 & 0.0 & 0.1 & 0.0 & 0.0 & 0.0 & 0.8 & 0.7 & 1.2 & 0.8 \\
				Zlib         & 0.0 & 0.0 & 0.0 & 0.0 & 0.0 & 0.0 & 0.0 & 0.0 & 0.1 & \textbf{0.2} & \textbf{0.2} & 0.2 & \textbf{1.2} & \textbf{0.9} & \textbf{1.3} & \textbf{1.1} \\
				Min-K\%      & 0.0 & 0.0 & 0.0 & 0.0 & 0.0 & 0.0 & 0.0 & 0.0 & \textbf{0.2} & 0.1 & 0.1 & 0.1 & 0.8 & 0.7 & 1.2 & 0.8 \\
				Min-K\%++    & 0.0 & 0.0 & 0.0 & 0.0 & 0.1 & 0.0 & \textbf{0.5} & 0.2 & \textbf{0.2} & 0.0 & 0.1 & 0.0 & 0.9 & 0.5 & 0.8 & 0.4 \\
				DC-PDD       & 0.0 & 0.0 & \textbf{0.2} & 0.0 & 0.0 & 0.0 & 0.0 & 0.0 & 0.0 & \textbf{0.2} & 0.1 & 0.0 & 0.5 & 0.1 & 0.1 & 0.2 \\
				Ref          & \underline{\textbf{0.2}} & \underline{\textbf{0.1}} & \underline{0.0} & \underline{\textbf{0.1}} & \underline{\textbf{0.2}} & 0.0 & \underline{0.3} & 0.0 & \underline{0.1} & 0.0 & 0.0 & 0.3 & 0.7 & 0.0 & 0.1 & 0.1 \\
				\rowcolor{RowGray}
				InfoRMIA1   & 0.1 & 0.0 & \underline{0.0} & 0.0 & 0.1 & \underline{\textbf{0.1}} & \underline{0.3} & \underline{\textbf{0.5}} & 0.0 & 0.0 & 0.0 & 0.0 & 0.1 & \underline{0.1} & \underline{0.2} & \underline{0.2} \\
				\rowcolor{RowGray}
				InfoRMIA2   & 0.0 & 0.0 & \underline{0.0} & 0.0 & 0.0 & 0.0 & 0.0 & 0.0 & \underline{0.1} & \underline{\textbf{0.2}} & \underline{\textbf{0.2}} & \underline{\textbf{0.4}} & \underline{0.7} & 0.0 & \underline{0.2} & 0.1 \\
				\bottomrule
			\end{tabular}%
		}
	\end{center}
\end{table}

\begin{table}[ht!]
\caption{AUC results on MIMIR benchmark with deduped Pythia models when using the first step checkpoint of the target model as the reference.}
\label{table:mimir_same_ref}
\begin{center}
\scriptsize
\setlength{\tabcolsep}{2pt}
\renewcommand{\arraystretch}{1.15}

\resizebox{\textwidth}{!}{%
\begin{tabular}{l *{16}{c}}
\toprule
& \multicolumn{4}{c}{\textbf{Wikipedia}} & \multicolumn{4}{c}{\textbf{Github}} & \multicolumn{4}{c}{\textbf{Pile CC}} & \multicolumn{4}{c}{\textbf{PubMed Central}} \\
\cmidrule(lr){2-5} \cmidrule(lr){6-9} \cmidrule(lr){10-13} \cmidrule(lr){14-17}
\textbf{Method}& 160M & 1.4B & 2.8B & 6.9B & 160M & 1.4B & 2.8B & 6.9B & 160M & 1.4B & 2.8B & 6.9B & 160M & 1.4B & 2.8B & 6.9B \\
\midrule
Loss         & 50.2 & 51.0 & 51.7 & 51.6 & 63.7 & 65.8 & 71.2 & 67.6 & 49.5 & 50.1 & 50.1 & 51.3 & 49.9 & 49.8 & 49.9 & 50.5 \\
Zlib         & \textbf{51.0} & 51.8 & 52.4 & 52.3 & \textbf{65.6} & \textbf{67.2} & \textbf{72.2} & 68.8 & 49.6 & 50.2 & 50.3 & 51.2 & 50.0 & 50.0 & 50.0 & 50.6 \\
Min-K\%      & 48.6 & 50.6 & 51.6 & 51.4 & 63.6 & 65.9 & 71.4 & 68.0 & 50.0 & 51.0 & 50.5 & 51.9 & 50.4 & 50.2 & 50.4 & 51.0 \\
Min-K\%++    & 47.7 & \textbf{52.3} & \textbf{53.7} & \textbf{52.4} & 61.4 & 65.7 & 70.7 & \textbf{69.1} & 49.8 & 51.1 & 49.9 & 51.7 & 50.9 & 50.6 & \textbf{51.2} & \textbf{52.3} \\
DC-PDD       & 49.0 & 50.6 & 52.4 & 51.8 & 64.9 & 66.2 & 71.4 & 69.0 & 49.6 & \textbf{51.1} & \textbf{51.2} & \textbf{51.9} & 50.5 & \textbf{51.0} & 50.6 & 51.1 \\
Ref          & 50.0 & \underline{50.9} & 51.6 & \underline{52.1} & 63.9 & 65.4 & \underline{71.4} & 66.9 & 49.4 & 50.0 & 50.2 & 51.3 & 49.8 & 50.0 & 49.5 & \underline{50.5} \\
\rowcolor{RowGray}
InfoRMIA1   & \underline{50.9} & 50.8 & 51.1 & 51.5 & \underline{65.0} & \underline{66.0} & 70.9 & \underline{66.9} & 49.4 & 49.5 & 49.8 & 50.5 & 50.2 & 49.8 & 49.4 & 49.8 \\
\rowcolor{RowGray}
InfoRMIA2   & 50.0 & 50.4 & \underline{51.7} & 51.6 & 63.5 & 65.3 & 70.5 & 66.9 & \underline{\textbf{50.6}} & \underline{50.9} & \underline{51.0} & \underline{51.4} & \underline{\textbf{51.4}} & \underline{50.3} & \underline{50.1} & 50.2 \\

\toprule
& \multicolumn{4}{c}{\textbf{ArXiv}} & \multicolumn{4}{c}{\textbf{DM Mathematics}} & \multicolumn{4}{c}{\textbf{HackerNews}} & \multicolumn{4}{c}{\textbf{Average}} \\
\cmidrule(lr){2-5} \cmidrule(lr){6-9} \cmidrule(lr){10-13} \cmidrule(lr){14-17}
\textbf{Method}& 160M & 1.4B & 2.8B & 6.9B & 160M & 1.4B & 2.8B & 6.9B & 160M & 1.4B & 2.8B & 6.9B & 160M & 1.4B & 2.8B & 6.9B \\
\midrule

Loss         & 50.7 & 51.4 & 51.9 & 52.5 & 49.0 & 48.6 & 48.3 & 48.4 & 49.2 & 50.4 & 51.2 & 51.7 & 51.8 & 52.4 & 53.5 & 53.4 \\
Zlib         & 50.0 & 50.8 & 51.3 & 51.8 & 48.2 & 48.1 & 48.0 & 48.1 & 49.6 & 50.2 & 50.9 & 51.0 & 52.0 & 52.6 & 53.6 & 53.4 \\
Min-K\%      & 50.0 & 51.2 & 52.2 & 52.7 & 49.4 & 49.3 & 49.1 & 49.3 & 50.2 & 51.3 & 52.4 & 53.0 & 51.7 & 52.8 & 53.9 & 53.9 \\
Min-K\%++    & 48.7 & 51.2 & \textbf{53.1} & 52.8 & \textbf{49.9} & \textbf{50.0} & \textbf{50.3} & \textbf{50.2} & \textbf{50.9} & 51.1 & 52.3 & 53.7 & 51.3 & \textbf{53.1} & 54.4 & \textbf{54.6} \\
DC-PDD       & 50.4 & \textbf{52.0} & 52.9 & \textbf{52.9} & 49.0 & 49.3 & 49.8 & 49.7 & 50.7 & 51.8 & \textbf{53.0} & \textbf{53.9} & 52.0 & 53.1 & \textbf{54.5} & 54.3 \\
Ref          & 50.3 & \underline{51.3} & \underline{51.8} & \underline{52.3} & 48.8 & 49.0 & 48.7 & 48.2 & 49.1 & 50.5 & 51.4 & 51.5 & 51.6 & 52.5 & 53.5 & 53.3 \\
\rowcolor{RowGray}
InfoRMIA1   & 50.3 & 51.2 & 51.2 & 51.6 & 48.0 & 47.7 & 47.8 & 47.8 & 50.4 & 50.7 & 51.1 & 51.2 & 52.0 & 52.3 & 53.0 & 52.8 \\
\rowcolor{RowGray}
InfoRMIA2   & \underline{\textbf{50.8}} & 51.0 & 51.6 & 52.2 & \underline{49.0} & \underline{49.1} & \underline{49.0} & \underline{48.8} & \underline{50.4} & \underline{\textbf{52.0}} & \underline{52.3} & \underline{52.3} & \underline{\textbf{52.2}} & \underline{52.7} & \underline{53.7} & \underline{53.3} \\
\bottomrule
\end{tabular}%
}
\end{center}
\end{table}

\section{Token-Based Analysis}
In this section, we provide some analytical results on the token-based interface on finetuned models on AG News and ai4privacy, when conducting offline Toekn-level InfoRMIA with 4 reference models.

\begin{figure}[!htb]
    \centering
    \includegraphics[width=0.5\linewidth]{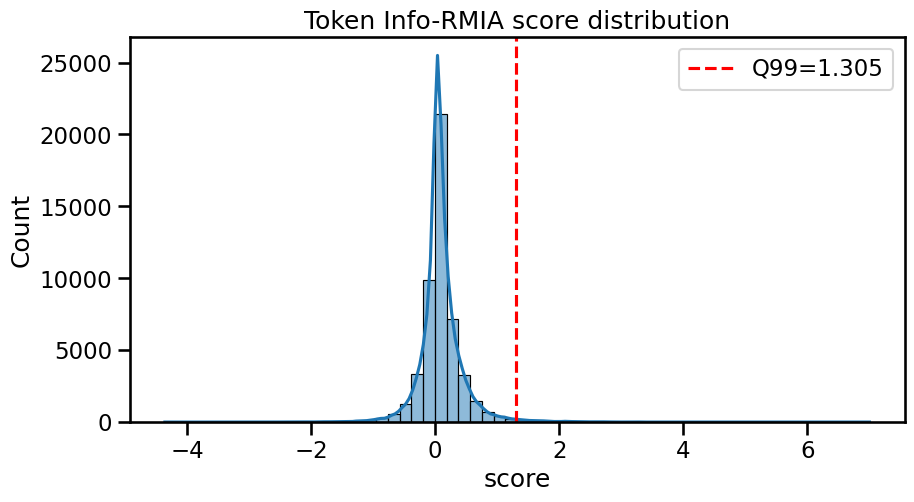}
    \caption{Distribution of token InfoRMIA scores on AG News dataset.}
    \label{fig:agnews_token_dist}
\end{figure}

\begin{table}[ht!]
\caption{Summary statistics of token membership scores grouped by entity type on AG News, sorted by mean scores. Top-$1\%$ scoring tokens are called "high" tokens. "n\_high" is the number of high tokens, and high\_rate" is the proportion of the tokens in each entity being high scoring. "None" entities are not nouns, which are unsurprisingly the majority.}
\label{table:agnews_entity_stats}
\begin{center}
\resizebox{\textwidth}{!}{%
\begin{tabular}{lccccccc}
\toprule
\bf entity & \bf count & \bf mean\_score & \bf median\_score & \bf p95 & \bf n\_high & \bf high\_rate \\
\midrule
PERSON      & 2225  & 0.156000  & 0.103894  & 0.847365  & 50  & 0.022472 \\
WORK\_OF\_ART & 107   & 0.135550  & 0.056464  & 0.695698  & 3   & 0.028037 \\
PRODUCT     & 75    & 0.122673  & 0.068920  & 0.854701  & 0   & 0.000000 \\
FAC         & 136   & 0.119761  & 0.067262  & 0.894820  & 1   & 0.007353 \\
LOC         & 161   & 0.117694  & 0.078282  & 0.729102  & 1   & 0.006211 \\
TIME        & 159   & 0.115637  & 0.080404  & 0.621918  & 1   & 0.006289 \\
ORG         & 4624  & 0.113697  & 0.073791  & 0.729737  & 74  & 0.016003 \\
GPE         & 1587  & 0.107486  & 0.064042  & 0.634189  & 20  & 0.012602 \\
QUANTITY    & 79    & 0.107218  & 0.076787  & 0.625024  & 1   & 0.012658 \\
MONEY       & 1139  & 0.103070  & 0.081028  & 0.464732  & 5   & 0.004390 \\
None         & 36188 & 0.094550  & 0.056949  & 0.646802  & 327 & 0.009036 \\
EVENT       & 188   & 0.094198  & 0.051614  & 0.684139  & 2   & 0.010638 \\
NORP        & 391   & 0.086809  & 0.063955  & 0.619991  & 4   & 0.010230 \\
ORDINAL     & 134   & 0.081780  & 0.039014  & 0.571259  & 0   & 0.000000 \\
CARDINAL    & 720   & 0.072828  & 0.051281  & 0.558225  & 4   & 0.005556 \\
PERCENT     & 123   & 0.052848  & 0.037016  & 0.407925  & 0   & 0.000000 \\
DATE        & 1798  & 0.048768  & 0.031457  & 0.501247  & 6   & 0.003337 \\
LAW         & 5     & 0.005115  & -0.096963 & 0.653592  & 0   & 0.000000 \\
LANGUAGE    & 3     & -0.149360 & -0.130233 & 0.017094  & 0   & 0.000000 \\
\bottomrule
\end{tabular}%
}
\end{center}
\end{table}

\begin{table}[ht!]
\caption{Summary statistics of token membership scores grouped by their private/non-private status in the ai4privacy dataset.}
\label{table:ai4privacy_stats}
\begin{center}
\resizebox{\textwidth}{!}{%
\begin{tabular}{lcccccccc}
\toprule
\bf token & \bf count & \bf mean & \bf std & \bf min & \bf 10\% & \bf 50\% & \bf 90\% & \bf max \\
\midrule
Non-private & 147411.0 & 0.090224 & 0.303371 & -4.658639 & -0.088854 & 0.056127 & 0.304669 & 8.894643 \\
Private     &  36340.0 & 0.076426 & 0.320213 & -4.478451 & -0.164783 & 0.048231 & 0.340272 & 7.925481 \\
\bottomrule
\end{tabular}%
}
\end{center}
\end{table}

\begin{figure}[!htb]
    \centering
    \includegraphics[width=0.9\linewidth]{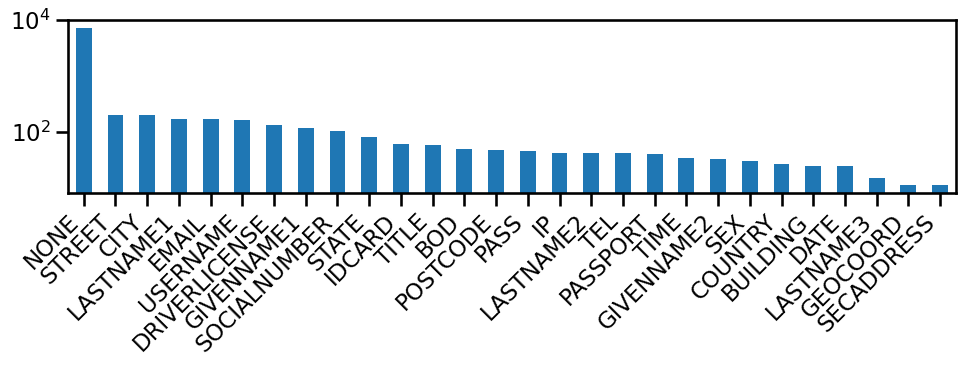}
    \caption{Distribution of high scoring tokens according to their types in ai4privacy dataset. The y-axis is the number of tokens in log scale.}
    \label{fig:high_token_ai4privacy_bar}
\end{figure}

\begin{figure}
    \centering
    \includegraphics[width=1\linewidth]{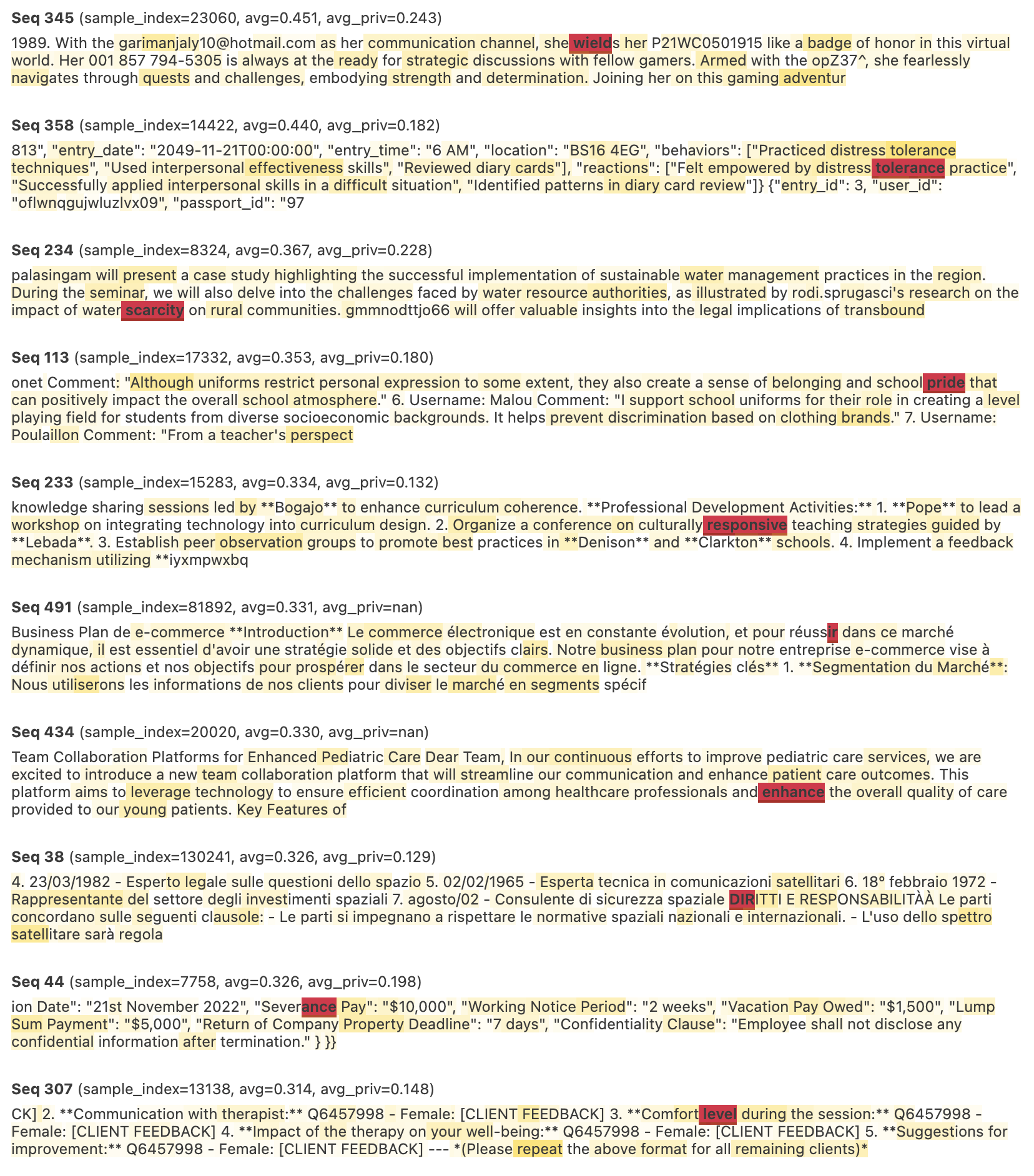}
    \caption{Top-10 memorized sequences in the ai4privacy dataset, ranked by sequence-based membership scores. Some of them do not even have any private tokens. Others have disproportionately small private token average scores.}
    \label{fig:top_10_seq}
\end{figure}

\begin{figure}
    \centering
    \includegraphics[width=1\linewidth]{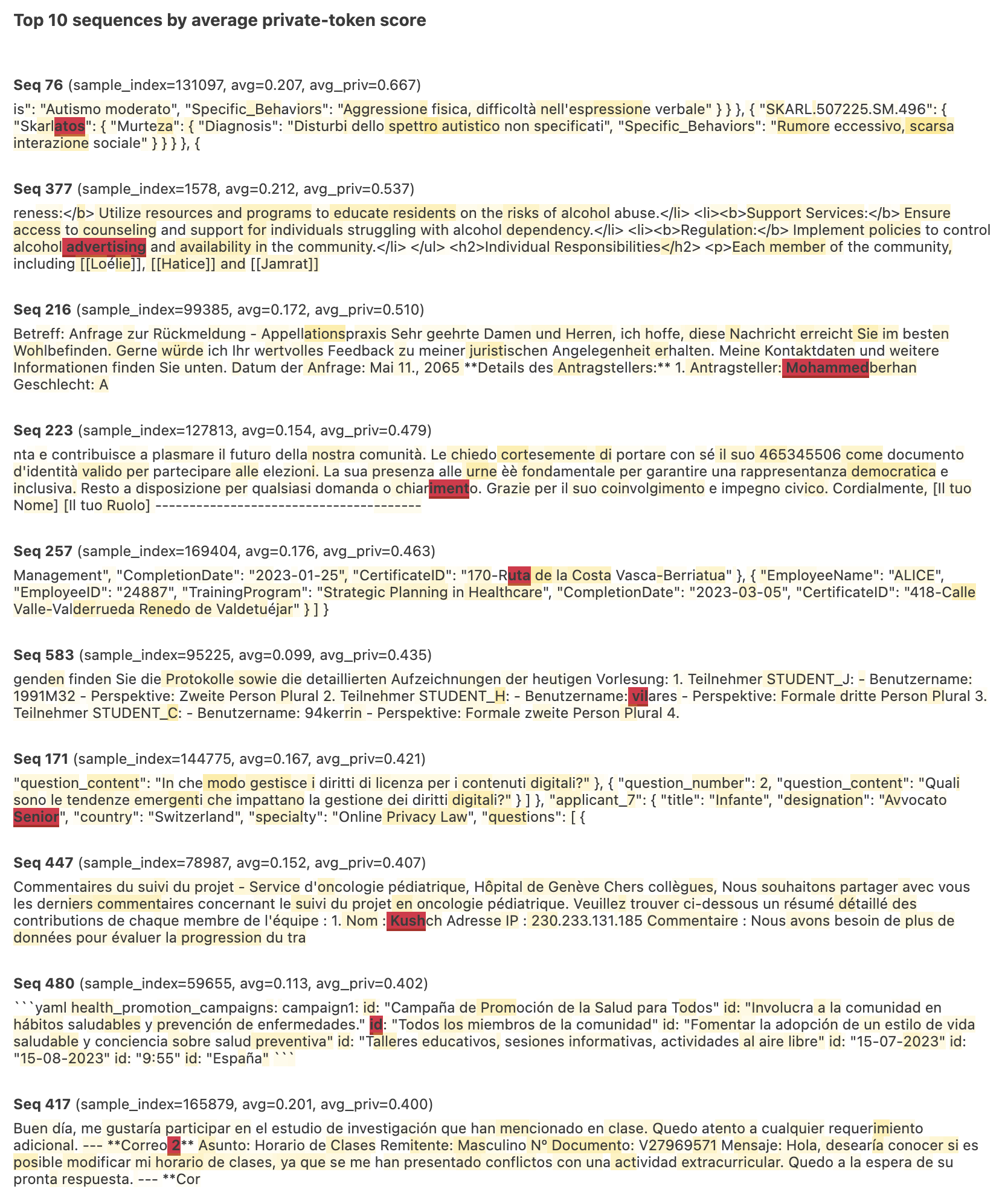}
    \caption{Top-10 sequences that have the highest average scores of private tokens. Note that their sequence averages are much smaller compared to those in Figure~\ref {fig:top_10_seq}, and the average private token scores. This is aligned with our intuition that signals from private tokens can get diluted in long texts.}
    \label{fig:top_10_token}
\end{figure}

\end{document}